\documentclass[3p]{elsarticle}

\usepackage[utf8]{inputenc}
\usepackage[english]{babel}
\usepackage[normalem]{ulem}
\useunder{\uline}{\ul}{}

\usepackage{textgreek}

\usepackage{hyperref}
\hypersetup{
    colorlinks=true,
    linkcolor=blue,
    filecolor=magenta,      
    urlcolor=cyan,
}

\usepackage[caption = false]{subfig}
\usepackage{multirow}
\usepackage{siunitx}
\usepackage[normalem]{ulem}
\useunder{\uline}{\ul}{}
\journal{Journal of Loss Prevention in the Process Industries}

\usepackage{array}
\usepackage{color}
\usepackage{colortbl}
\usepackage{multirow}
\usepackage{pifont}

\begin{document}

\begin{frontmatter}

\title{Experimental Large-Scale Jet Flames' Geometrical Features Extraction for Risk Management Using Infrared Images and Deep Learning Segmentation Methods}

\author[addr1]{Carmina Pérez-Guerrero}

\author[addr2]{Adriana Palacios \corref{cor1}}

\author[addr1]{Gilberto Ochoa-Ruiz \corref{cor1}}

\author[addr3]{Christian Mata}

\author[addr3]{Joaquim Casal}

\author[addr1]{Miguel Gonzalez-Mendoza}

\author[addr1]{Luis Eduardo Falcón-Morales}

\cortext[cor1]{
Corresponding author: adriana.palacios@udlap.mx; gilberto.ochoa@tec.mx
}

\address[addr1]{Tecnologico de Monterrey, School of Engineering and Sciences, Jalisco, 45138, Mexico.}
\address[addr2]{Universidad de las Americas Puebla, Department of Chemical, Food and Environmental Engineering, Puebla, 72810, Mexico.}
\address[addr3]{Universitat Politècnica de Catalunya. EEBE, Eduard Maristany 16, 08019 Barcelona. Catalonia, Spain.}

\begin{abstract}
Jet fires are relatively small and have the least severe effects among the diverse fire accidents that can occur in industrial plants; however, they are usually involved in a process known as the domino effect, that leads to more severe events, such as explosions or the initiation of another fire, making the analysis of such fires an important part of risk analysis. This research work explores the application of deep learning models in an alternative approach that uses the semantic segmentation of jet fires flames to extract the flame’s main geometrical attributes, relevant for fire risk assessments. A comparison is made between traditional image processing methods and some state-of-the-art deep learning models. It is found that the best approach is a deep learning architecture known as UNet, along with its two improvements, Attention UNet and UNet++. The models are then used to segment a group of vertical jet flames of varying pipe outlet diameters to extract their main geometrical characteristics. Attention UNet obtained the best general performance in the approximation of both height and area of the flames, while also showing a statistically significant difference between it and UNet++. UNet obtained the best overall performance for the approximation of the lift-off distances; however, there is not enough data to prove a statistically significant difference between Attention UNet and UNet++. The only instance where UNet++ outperformed the other models, was while obtaining the lift-off distances of the jet flames with 0.01275 m pipe outlet diameter. In general, the explored models show good agreement between the experimental and predicted values for relatively large turbulent propane jet flames, released in sonic and subsonic regimes; thus, making these radiation zones segmentation models, a suitable approach for different jet flame risk management scenarios.
\end{abstract}

\begin{keyword}
jet fire; risk management; deep learning; semantic segmentation; flame height; lift-off.
\end{keyword}

\end{frontmatter}

%%\linenumbers

\section{Introduction}
Certain industrial activities, that take place in process plants or during the storage and transportation of hazardous materials, can face severe accidents. The effects of these major accidents can have consequences beyond the activity’s boundary and affect external factors such as human health, environmental impact, or property damage. The accuracy in the evaluation of the effects and consequences that derive from these accidents presents a great area of opportunity that can be tackled by the improvement of the knowledge of the main features of the most common major accidents. A complete understanding of these features is essential to avoid them, or in the worst-case scenario, to reduce and limit their consequences and severity. Some accidents are well known and have been thoroughly explored nowadays, however, some other accidents present various gaps in knowledge that could be investigated.

Jet fires are one kind of fire accident that have been repeatedly found to initiate domino effects, whose consequences could lead to catastrophic events in the process industries.  Gómez-Mares, Zárate, and Casal \cite{Gomez08}  performed a historical survey based on over 84 cases obtained from four European accident databases. They reported that one in two jet fire events caused a domino effect, originating at least another fire accident with severe effects; explosions happened in 56\% of the cases, a vapor cloud was generated in 26\% of the cases, and other types of fires occurred in 27\% of the cases; these percentages do not add up to 100\%, given that more than one of these events can occur at the same time.  It was also found that propane was the most common fuel involved since jet fires are usually originated by the loss of containment and ignition of flammable gas, vapor, or spray. The flammable elements may be released through a hole, a broken pipe, a flange, or during process flares.

Jet fires tend to occur in rather compact spaces and the probability of flame impingement on another equipment from the associated domino effect is rather high. Therefore, even though jet fires tend to be confined to relatively shorter distances compared to other fire accidents, such as pool fires, flash fires, or fireballs, they have some specific features that can significantly increase the hazards associated with them. These features make them very important from a risk analysis point of view, which leads to the significance of their proper characterization. Several authors have investigated the main geometrical features of jet flames. Some of these studies have been focused on the jet flame height (\cite{Gautam84}, \cite{Bradley16}, \cite{ Guiberti20}) with recently proposed equations to predict jet flame height including several fuels and orientations \cite{Bradley16}. The jet flame shape has also been predicted through several models (\cite{palacios11}, \cite{ZHANG15}, \cite{WANG21}), as well as the non-ignition zones (\cite{Bradley16}, \cite{Wang19}).

The most common jet fire consequence analyses rely mostly on empirical models that use a simplified flame geometry and fixed emissive power to determine the thermal radiation load on a target, thus, a popular approach to evaluate both jet fire geometric features and radiation loads has been the use of Computational Fluid Dynamics (CFD) \cite{colella2020}. However, Mashhadimoslem et al. \cite{mashhadimoslem20} have recently found that CFD methods require more computational time and have a higher cost impact than neural networks; less computation time and higher accuracy are the hallmarks of neural networks for predicting processes, as well as their ability to work on more realistic data. 

In the present work, the benefits of neural networks, from the field of Deep Learning, are assessed as a tool to develop an alternative approach that uses image segmentation for the approximation of jet flame height, area, and non-ignition flame zone. These are relevant geometric features used in jet fire risk analysis, more specifically, to control the likelihood of direct flame impingement and to determine the distribution and intensity of radiant heat that is emitted from the flame to the surrounding equipment. Furthermore, the explored image segmentation approach can be used to optimize the parameters of other radiation and turbulence models, making the characterization of such geometrical and radiation factors, critical for the process safety field. In the present work, experimental data on large-scale jet flames, released in both sonic and subsonic conditions, have been used along with recent developments in the image processing and computer vision fields, namely, recent algorithmic strides in the form of deep learning methods, to produce highly accurate segmentation masks used in image analysis tasks that characterize jet fires.

Deep learning algorithms, and more specifically, Convolutional Neural Networks (CNNs), are among the most extensively used and increasingly influential types of machine learning algorithms for computer vision tasks and data-driven models. Convolutional networks have shown outstanding capabilities in many complex endeavors, such as image recognition, object detection, and semantic and instance segmentation \cite{Litjens_2017}. Advancements on these techniques have been proliferating, and with cutting-edge research being published constantly, better and more robust algorithms are emerging faster than ever.

\begin{figure}[!htbp]
\centering
\subfloat[]{\includegraphics[width = 3in]{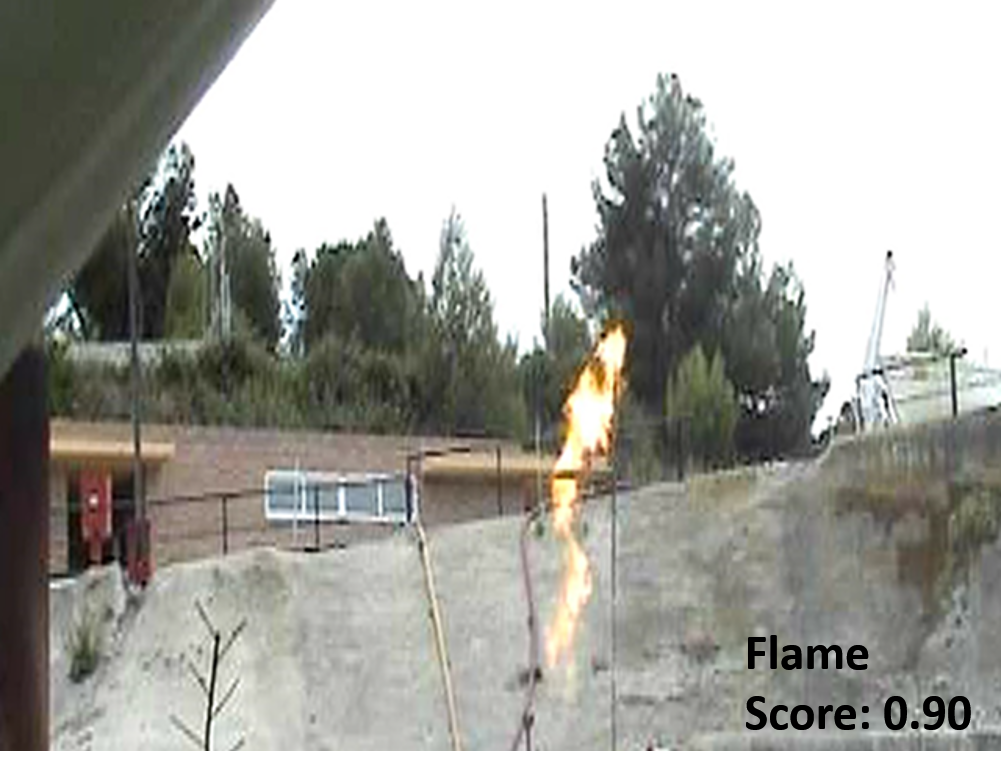}}
\qquad
\subfloat[]{\includegraphics[width = 3in]{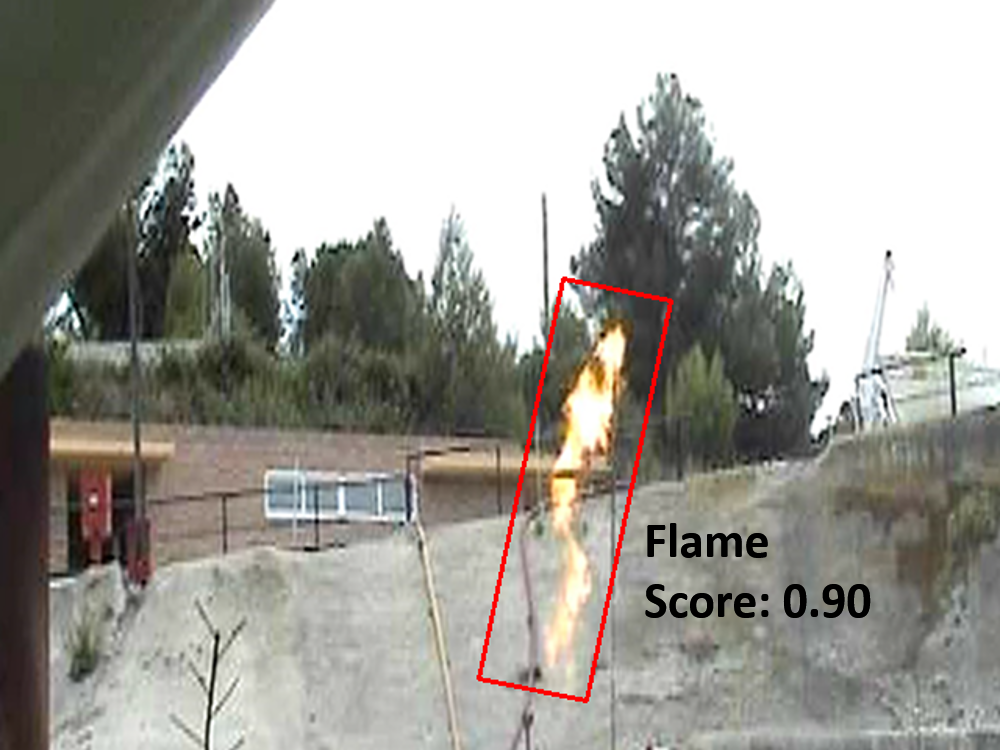}}

\subfloat[]{\includegraphics[width = 3in]{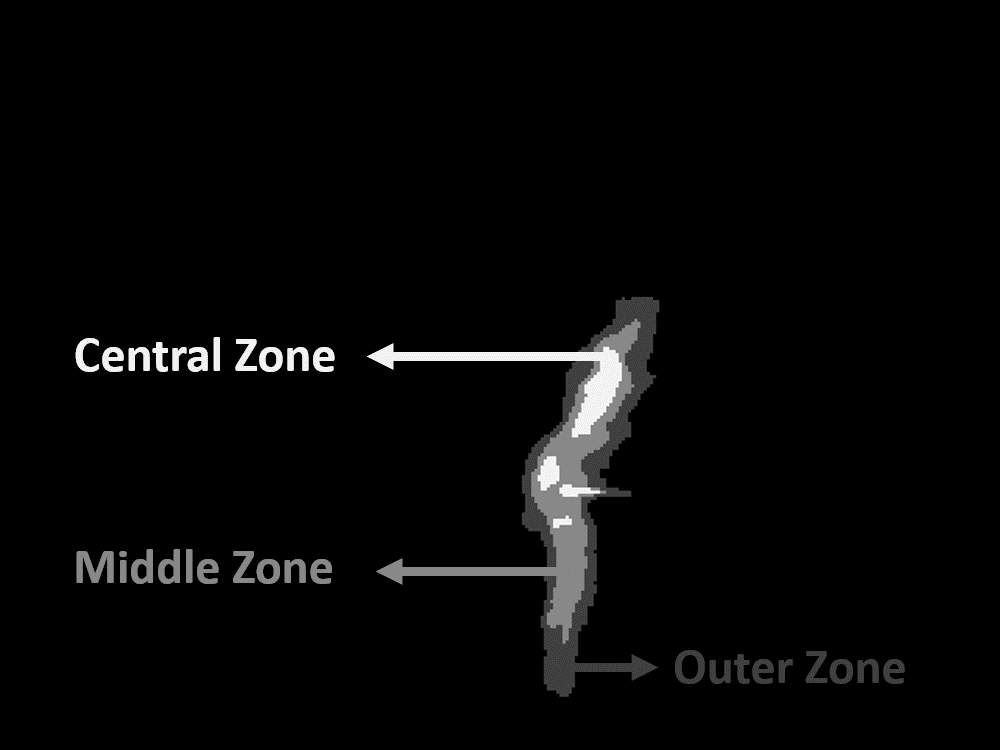}}
\qquad
\subfloat[]{\includegraphics[width = 3in]{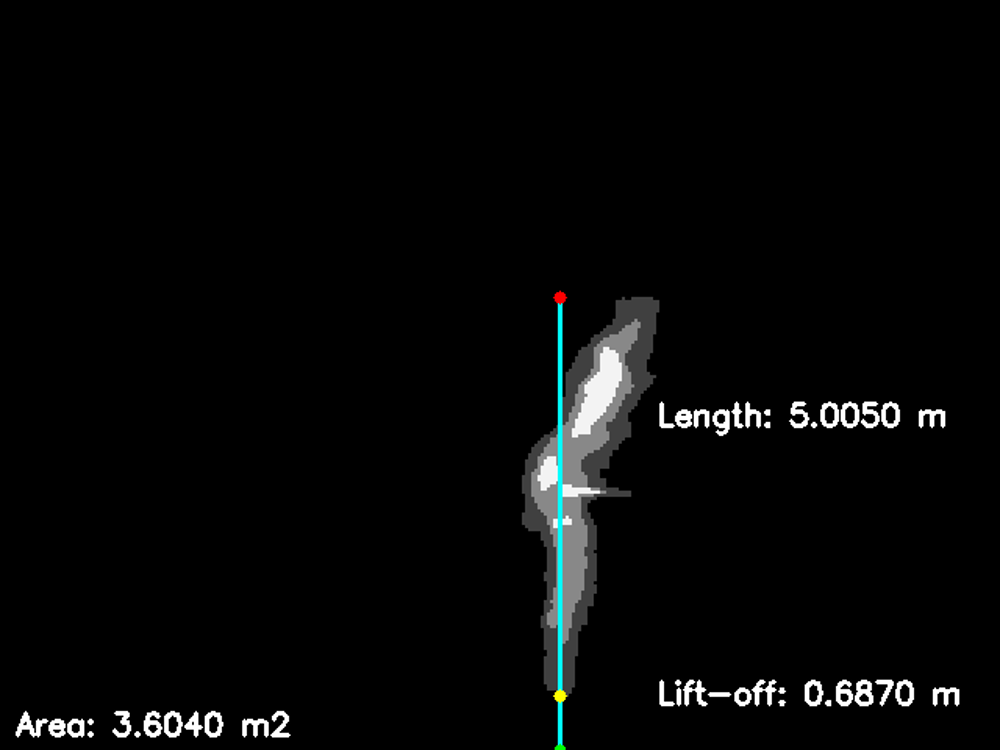}}
\caption{Different computer vision tasks performed on a vertical jet flame of propane. Figure (a) is an example of object classification denoting the presence of the flame in the image. Figure (b) is an example of generic object detection using a bounding box for the localization of the flame. Figure (c) is an example of semantic segmentation, signalling the name of each segment within the flame. Figure (d) is an example of the main geometrical characteristics that are extracted from the segmentation.}
\label{fig:cv}
\end{figure}

Among the tasks where convolutional networks excel, instance and semantic segmentation have seen a recent surge in technical progress. It extends and combines two of the most popular computer vision tasks: object detection and semantic segmentation. Its applications range from autonomous vehicles \cite{Chang_2021} to robotics \cite{Xie_2021} and surveillance systems \cite{Zhang_2020}, to name a few. Instance segmentation aims to assign class labels and pixel-wise instance masks to identify a varying number of instances presented in each image or video frame \cite{Liu_2018}. Figure \ref{fig:cv} illustrates the main differences between object classification, object detection, and semantic segmentation, along with an example of the main geometrical characteristics that are extracted from the segmentation. Object classification tries to assign a certain category to a full image, object detection focuses on locating and classifying an object with high precision, and semantic segmentation is interested in finding the pixels of the image that correspond to certain elements, which in the case of this research, is the boundaries of the jet fires.

Segmentation is a pre-processing stage from which other related problems can be solved.  It can be used as an attention mechanism by focusing only on relevant parts of the images that can be further analyzed, such as faces for bio-metric studies \cite{khan20} or in the medical image analysis domain for image-guided interventions, radiotherapy, or improved radiological diagnostics \cite{taghanaki20}. Segmentation is useful for fine-grained tasks where a bounding box around the objects is too coarse for a solution, especially when the scene is cluttered or when the precise entity boundaries are important \cite{hosseini20}. Segmentation can also be used to change the pixels of an image into higher-level representations that are more meaningful, obtaining segments that represent object textures, locations, or shapes, which is important, for example, in the automatic interpretation of remote sensing images \cite{troya16}.

This paper uses the information obtained from exploratory research on different semantic segmentation approaches, and a selection process of evaluation metrics, to single out the best method for the segmentation of fire ration zones. The selected model is trained on horizontal jet fires and then used to perform image segmentation on vertical jet fires. As a practical application of this approach, the segmented images are used to extract the flame's geometrical information, and the results are then validated through previous characterization experiments.

Obtaining a model that can use the infrared features of the flame to extract its geometrical information is a promising first step for the development of a reliable risk management system that could be used for general fire incidents. The segmentation experiments presented in this work could later be used to characterize the different radiation zones to improve the prediction of thermal radiation load on a target, going beyond defining just the main jet fire geometric features, which is the current focus of this research.
Therefore, the work described in this manuscript is a contribution to the knowledge of the architectures that perform well for the segmentation of radiation zones within the fire, the most representative evaluation methods to an expert's criteria, and practical results obtained from the proposed approach, showing how it can be used in the prediction of the main geometrical features of jet fires for accurate risk assessment and better prevention and control of jet fire accidents, which have been the origin of severe accidental scenarios, occurring worldwide in industrial establishments and during the transportation of hazardous materials.

The rest of this paper is organized as follows: Section \ref{sec:literature} describes the previous work found in the literature related to the problem. Section \ref{sec:materials} contains a general description of the jet fire experimental set-up and their characteristics, followed by a brief description of the data set used to train segmentation algorithms and architectures. Section \ref{sec:methodology} describes the different segmentation approaches and introduces the definition of the flame's temperature boundaries and geometrical characteristics. This section also explains the data pre-processing methods applied to the data set, the evaluation metrics used to compare and validate the models, and finally, the training process established for the training of the segmentation algorithms. Section \ref{sec:results} presents the results of the comparison between the different segmentation models, along with the evaluation of the geometrical information extracted from the segmentation obtained through the best performing models. Finally, Section \ref{sec:conclusions} summarizes the major findings of the presented research and offers a discussion regarding the future work and areas of opportunity.

\section{Literature Review}
\label{sec:literature}
\subsection{Jet Fire Thermal Radiation}
Thermal radiation frequently plays a major role in the heat transfer process, therefore, fast and reliable tools to compute radiation emanating from flames are necessary for risk engineering applications. Some models are based on algebraic expressions and assume an idealized flame shape with uniform surface radiation emissive power, such as a cylindrical crone, proposed by Croce and Mudan \cite{croce1986}, or a frustum of a cone proposed by Chamberlain \cite{Chamberlain87}.

Some other models, such as the line source emitter model, assume that the radiation source is represented as a line at the center of the flame. This model was used by Zhou and Jiang, who also proposed a profile of emissive power per line length that can predict the radiant heat flux of jet flames with different flame shape simulations \cite{zhou15}.

Other models assume that the radiation is emitted by a single or multiple point sources distributed along the flame surface. The Single Point Source model relies on the assumption that the radiation intensity emitted from the source is isotropic, but on the other hand, the Weighted Multi Point Source model considers that the radiation emanates from N point sources distributed along the main axis of the flame. In Quezada, Pagot, Fran\c{c}a, and Pereira \cite{quezada2020}, they propose a new version of the inverse method for optimizing the Weighted Multi Point Source model parameters. The proposed new optimized weighting coefficients and the new correlation for the radiative heat transfer rate showed improved results for predicting the radiative heat transfer from the experiments.

A different approach to evaluate the consequences of the industrial hazards that jet fire accidents represent, is through the use of Computational Fluid Dynamics (CFD). The models mentioned previously lack consideration of the geometry of the system, which may make them unreliable when dealing with barriers or equipment, therefore CFD represent an alternative solution for evaluating both the jet fire’s geometric features and thermal radiation. For instance, Kashi and Bahoosh \cite{Kashi2020} explored different radiation and turbulence models along with CFD simulations to evaluate how firewalls or equipment around the flame can affect the radiation distribution and temperature profile of the jet fire.

\subsection{Jet Fire Geometrical Features}
Fires are one of the most serious threats to the safety of both the equipment and the people who operate in process units. However, it is possible to identify the optimal space for the location of equipment and structures, by evaluating the size and shape of the flames. Therefore, being able to predict the shape and proportions of a jet fire is extremely important from a risk management point of view \cite{Kashi2020}. Previous research has been done to characterize the main geometrical aspects of jet flames. For instance, Gautam \cite{Gautam84} presents a systematic study of the factors that affect the lift-off height, which is the distance between the burner exit and the base of the lifted flame. It is found that the lift-off height varies linearly according to the exit velocity, and it is independent of the burner diameter.

Palacios and Casal \cite{palacios11} analyzed the flame shape, length, and width of relatively large vertical jet fires at sonic and subsonic exit velocities. The flame boundary was defined at a temperature of 800 K, and the results indicated that a cylindrical shape could accurately describe the shape, length, and diameter of the flame.

Zhang et al. \cite{ZHANG15} investigate the flame shape and volume of turbulent gaseous fuel jets with different nozzle diameters. Baron's model and Orloff's image analysis method for shape prediction are compared. The results show that Baron’s method offers good predictions of the width and shape of the flame, with the width predictions being only slightly larger than the experimental results at the bottom and top sections of the flame. Based on the results, a mathematical model for flame volume estimation is deduced by the integration of the Baron’s expression.

Bradley, Gaskell, Gu, and Palacios \cite{Bradley16} introduce an extensive review and re-thinking of jet flame heights and structure, including a proposal of dimensionless correlations for the atmospheric jet flames' heights, lift-off distances, and mean flame surface densities. It was found that the same flow rate parameter could be used to correlate both heights and flame lift-off distances, and based on that, an equation was proposed to predict jet flame height including several fuels and orientations.

Guiberti, Boyette, and Roberts \cite{Guiberti20} explore the Delitcharsios' model to obtain the flame height for subsonic jet flames at elevated pressure. The results show that the Delitcharsios' model predicts well around 20\% of the flame height in these cases, so a range of two empirical constants are suggested to improve the predictions of the flame height equation.

Wang et al. \cite{Wang19} introduce a systematic experimental study on the lift-off behavior of jet flames impinging on cylindrical surfaces. The experiments are comprised of varying fuel exit velocity, internal diameters of the nozzle, and nozzle-to-surface spacing. An image visualization technique was used to reconstruct the image frames of a camera and accurately calculate the flame lift-off distance. It was observed that the lift-off distance depends on not only the exit velocity but also on the increase of nozzle diameter. It is also found that a dimensionless flow number expression for the lift-off distance provides the best fit.

Mashhadimoslem, Ghaemi, and Palacios \cite{mashhadimoslem20} proposed an Artificial Neural Networks (ANN) approach that uses the mass flow rates and the nozzle diameters to estimate the jet flame lengths and widths. The two methods explored were a Multi-layer Perceptron method with Bayesian Regularization back-propagation and a Radial Based Function method with a Gaussian function. The two methods show negligible discrepancies between them and can be used instead of Computational Fluid Dynamic methods, which require more computational time and have a higher cost impact than neural networks.

Wang et al. \cite{WANG21} present systematic experiments that show the effects of the nozzle exit velocity, diameter, and exit-plate spacing on the horizontal impingement of jet fires. The flame pattern and color are observed to evolve according to an increase in exit velocity and a decrease in exit-plate spacing. An Otsu method is used to calculate the flame intermittency probability. The probability contour of 50\% is used to determine the flame geometric features of interest to estimate the flame extension area. It is found that the temperature profile of the fire results in a big difference in the upward and downward directions along the vertical plate, so a uniform correlation, with the radius as the characteristic length scale, is proposed.

\subsection{Automatic Fire and Smoke Monitoring}

In the context of this research, segmentation is used to validate flame characterization experiments by extracting geometrical information that can be used to reinforce principles of technological risks and industrial safety. There has been previous work in the literature that have sought to apply automatic methods to perform risk assessments in industrial security applications. 

For instance, Janssen and Sepasian \cite{Janssen18} presented an automatic flare-stack-monitoring system with flare-size tracking and automatic event signaling using a computer vision approach where the flare is separated from the background by using temperature thresholding. To enhance the visualization of the temperature profile of the flare, false colors are added, representing the different temperature regions within the flare. Similarly, in Zhu et al. \cite{zhuli20} an infrared image-based feedback control system was proposed to aid in fire extinguishing through automatic aiming and continuous fire tracking. To identify and localize the fire, an improved maximum entropy threshold segmentation method was employed. Threshold segmentation is the simplest method of image segmentation and has a fast operation speed. However, it is not a very good method to identify regions, but it represents a first step if combined with other segmentation techniques. \cite{yuheng17}.

Gu, Zhan, and Qiao \cite{Gu20} proposed a Computer Vision-based Monitor of Flare Soot, namely VMFS, which is composed of three stages. First, they apply a broadly tuned color channel to localize the flame, then they perform a saliency detection through K-means to determine the flame area, and finally use the center of it to search for a potential soot region. K-means clustering provides an efficient shape-based image segmentation approach and has been used in other fire segmentation applications \cite{rudz13}, \cite{ajith19}. Nonetheless, K-means is not very effective when dealing with clusters of varying sizes and density since the centroids can be dragged by outliers \cite{zaitoun15}, and therefore the inherent dynamism of the fire may affect its results.

Ajith and Martínez-Ramón \cite{ajith19} proposed a computer vision-based fire and smoke segmentation system. Based on a feature extraction process, they test multiple clustering algorithms: K-Means clustering, Gaussian Mixture Model (GMM), Markov Random Fields (MRF), and Gaussian Markov Random Fields (GMRF). Even if MRF was found to be the model that distinguishes fire, smoke, and background in a more precise manner, GMM was able to segment most of the fire regions but yielded a much lower accuracy for smoke segmentation. Bayesian methods for clustering such as these have no free parameters to adjust; however, an assumption on the class of the likelihood functions and the prior probabilities is needed for them to perform well.

Roberts and Spencer \cite{roberts19} proposed a method that introduces a new fitting term to the Chan-Vese algorithm that allows the background to consist of multiple regions of inhomogeneity. This algorithm belongs to the group of active contour models and is the most successful one among them when dealing with infrared image segmentation \cite{Zhou16}. The Chan-Vese model can achieve sub-pixel accuracy of object boundaries and provides closed and smooth contours but it depends on several tuning parameters that have to be manually selected and is used primarily for binary labels, so it needs to be applied multiple times or generalized to obtain multiple segments.

These different approaches perform well for data sets that they were developed on, however, it is necessary to choose which features are important for each specific problem, so many aspects involve domain knowledge and long processes of trial and error. Deep Learning architectures such as CNNs, in contrast, are able to discover the underlying patterns of the images and automatically detect the most descriptive and salient features with respect to the problem \cite{Mahony19}.

\subsection{Deep Learning in Fire Scenarios}

Convolutional Neural Networks have achieved state-of-the-art results in many computer vision tasks and an improvement on these models, that can recognize images at the pixel level and ensure robustness and accuracy simultaneously, is found in Fully Convolutional Networks (FCNs) \cite{Mingwei20}. These architectures have the ability to make predictions on arbitrarily sized inputs and have demonstrated outstanding results for semantic image segmentation tasks; however, the propagation through several convolutional and pooling layers affects the resolution of the output feature maps. Some advanced FCN-based architectures have been proposed to address this flaw, some of the most notable are Deeplab, SegNet, and UNet.

DeepLabv3 \cite{chen18} is composed of two steps, the encoding phase, where information about the object's presence and location in an image is extracted, and the decoding phase, where the extracted information is then used to reconstruct an output of appropriate dimensions. This architecture recovers detailed structures that may be lost due to spatial in-variance and has wider receptive fields; however, it shows low accuracy on small-scaled objects and has trouble capturing delicate boundaries of objects.

Segnet \cite{Badrinarayanan2016} consists of an encoder network, a corresponding decoder network, and a pixel-wise classification layer. It is designed to be efficient both in terms of memory and computational time during inference but its performance drops with a large number of classes and is sensitive to class imbalance.

The UNet \cite{Ronneberger2015} architecture consists of a contracting path that captures the context of images and a symmetric expanding path that enables precise localization of the segments. To predict the pixels in the border region of the image, the missing context is extrapolated by mirroring the input image, which allows for precise localization of the regions of interest. This architecture, however, takes a significant amount of time to train since the size of the network is comparable to the size of the features. It can also leave a high GPU memory footprint when dealing with large images. The Attention UNet \cite{oktay18} is an improved version of UNet that uses attention gates to avoid excessive and redundant use of computational resources and feature maps; however, these mechanisms add more weight parameters to the model, which further increases the training time.

UNet++ \cite{zhou19} is another improvement on the UNet architecture, which implements redesigned skip connections that aggregate features of different semantic scales, across an efficient ensemble of U-Net models of varying depths that partially share an encoder. This improvement overcomes the downsides of UNet that involve the unknown optimal network depth and the restrictive fusion scheme of feature maps in its skip connections; however low-level encoder features may still be used repeatedly at multiple scales, increasing the computational resources needed.

Deep Learning methods have been proven to perform far better than traditional algorithms; however, they do present some drawbacks, mainly in terms of computing requirements and training time. When dealing with a limited training data set, the Deep Learning models may overfit and not be able to generalize for the task outside of the training data and it would be difficult to manually tweak the model because of their great number of parameters and their complex interrelationships, making these models a black box in comparison to the transparency of the traditional computer vision approaches \cite{Mahony19}.

\section{Materials}
\label{sec:materials}
\subsection{Experimental set-up}
\begin{figure}[!htbp]
\centering
\includegraphics[width=6in]{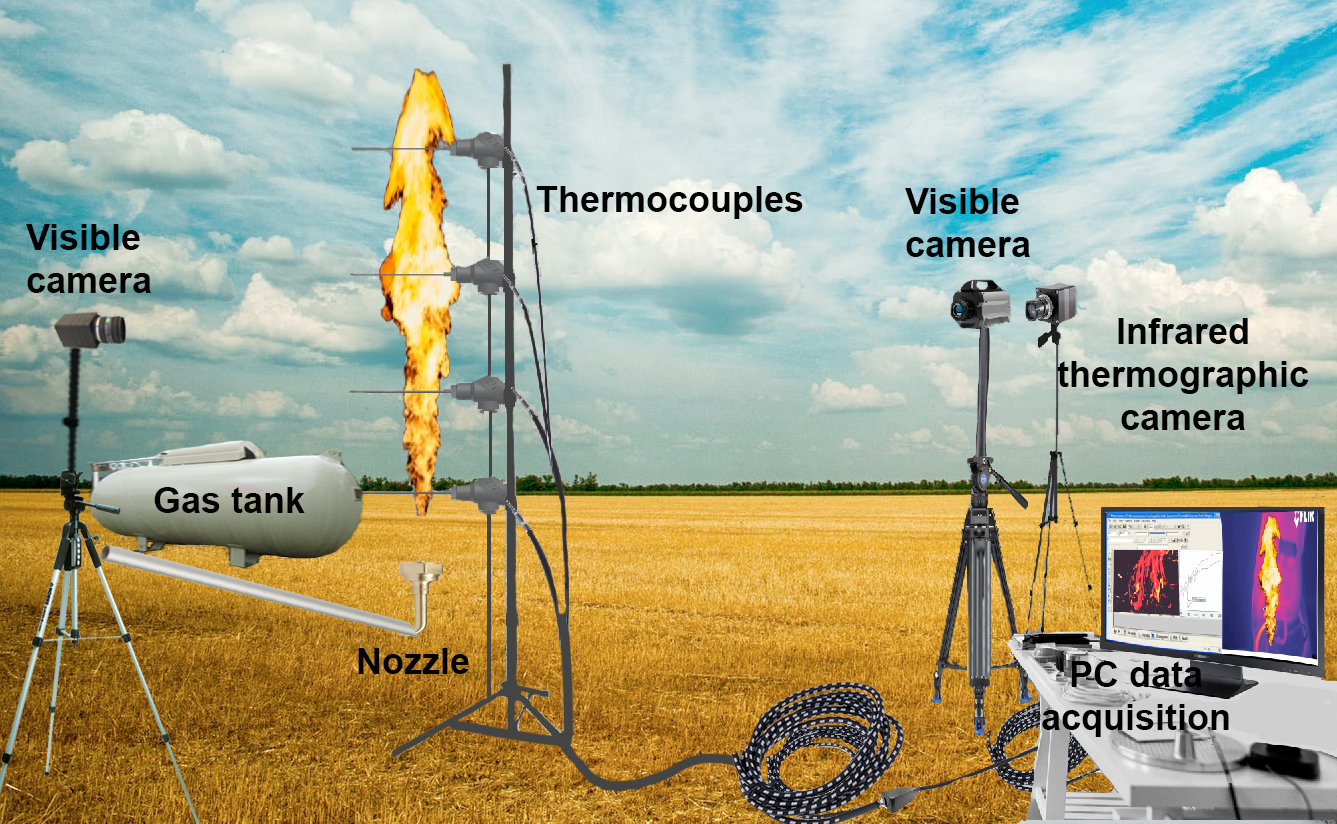}
\caption{A schematic view of the experimental set-up.}
\label{fig:setup}
\end{figure}

Jet fires, obtained from large-scale tests in the open field, have been analyzed in the present study. The characteristics of the utilized dataset are shown in Table 1. The jet ﬁre experiments involve subsonic and sonic vertical ﬂames of propane, discharged with nozzles ranging between 12.75 mm and 30 mm. The mass ﬂow rates ranged between 0.03 kg/s to 0.4 kg/s. The experiments were ﬁlmed with an infrared thermographic camera (Flir Systems, AGEMA 570) and two video cameras registering visible light (VHS). Four images per second were obtained from the infrared (IR) camera; while twenty ﬁve digital images per second were obtained from the video cameras, which resolutions were 384 x 288 pixels and 320 x 240, respectively. The two visible cameras were located orthogonally to the ﬂame, and one of them was located next to the infrared thermographic camera. The jet ﬂame axial temperature distribution was also measured using a set of thermocouples along the jet ﬂame centreline. The schema of the experimental set-up is shown in Figure \ref{fig:setup}. Further details of the experimental set-up can be found in \cite{Adriana11}.

\begin{table}[!htbp]
\centering
\caption{Experimental conditions of the large-scale jet ﬁre tests performed \cite{Adriana11}.}
\label{tab:features}
\begin{tabular}{cccccc}
\hline
\textbf{\begin{tabular}[c]{@{}c@{}}Experimental \\ Test\end{tabular}} & \textbf{\begin{tabular}[c]{@{}c@{}}Pipe outlet \\ diameter (m)\end{tabular}} & \textbf{\begin{tabular}[c]{@{}c@{}}Wind speed\\ (m/s)\end{tabular}} & \textbf{\begin{tabular}[c]{@{}c@{}}Ambient\\ temperature (°C)\end{tabular}} &
\textbf{\begin{tabular}[c]{@{}c@{}}Fuel velocity \\ (m/s)\end{tabular}} &
\textbf{\begin{tabular}[c]{@{}c@{}}Treated\\ images\end{tabular}} \\ \hline
1       & 0.01275       & 0.4       & 28        & 27-255     &129   \\ \hline
2       & 0.015         & 0.4       & 28        & 233-245    &50   \\ \hline
3       & 0.02          & 0.4       & 30        & 33-256     &79   \\ \hline
4       & 0.03          & 0.43      & 30        & 95-254     &42   \\ \hline
\end{tabular}
\end{table}

\subsection{Dataset}

\begin{figure}[!htbp]
\centering
\subfloat[]{\includegraphics[width = 3in]{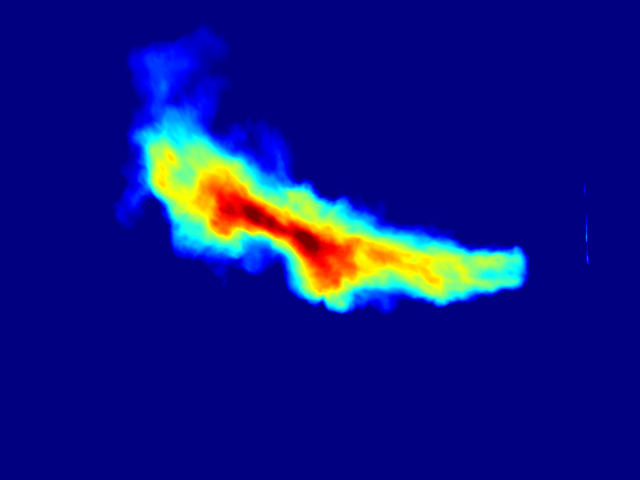}}
\qquad
\subfloat[]{\includegraphics[width = 3in]{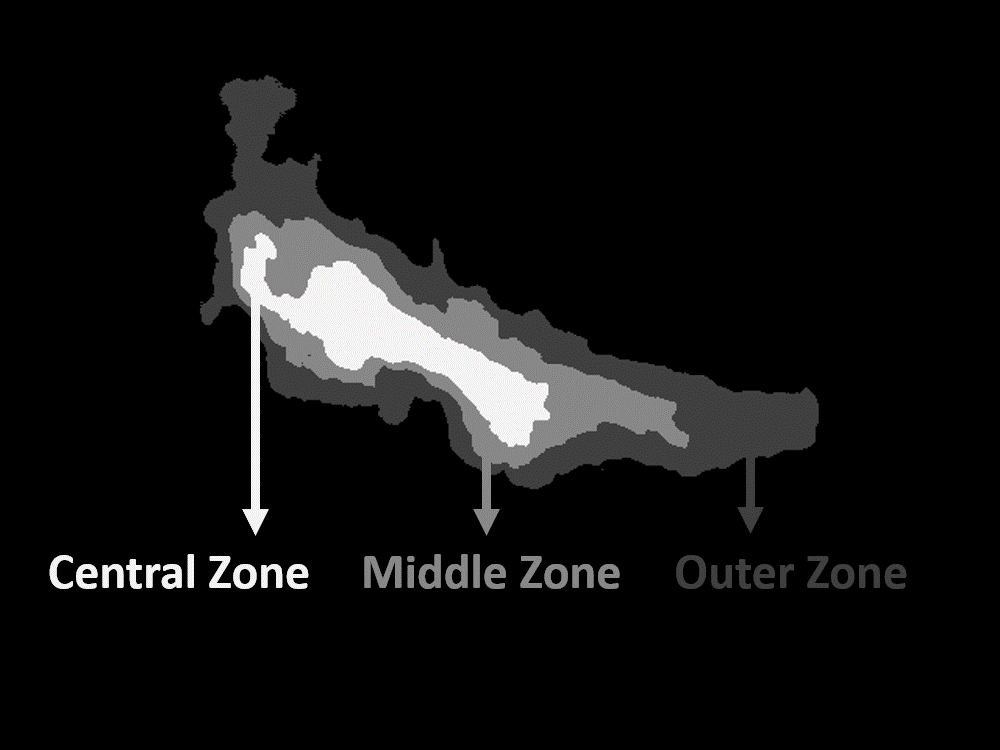}}
\caption{Sample segmentation of the radiation zones. Image (a) is an infrared visualization of an horizontal propane jet flame. Image (b) is the corresponding ground truth segmentation, with the segment names. Modified from \cite{Vahid21}.}
\label{fig:horizontal}
\end{figure}

To perform the experiments with Machine Learning and Deep Learning models, a data set of visual and infrared images of horizontal jet fires was employed. These flames were experimentally obtained at subsonic and sonic gas exit rates by three of the present authors. Each image has its respective temperature zones segmented and validated to serve as ground truth, as observed in Figure \ref{fig:horizontal}. In total, the data set contains 201 samples of paired infrared images and ground truth segmentation masks \cite{Vahid21}.

\section{Methodology}
\label{sec:methodology}
\subsection{Jet flame boundary and geometrical features}
\label{sec:boundary}
The vertical propane jet flame experiments, shown in Table 1, were analyzed to determine the fire temperature zones variation as a function of the jet flame axial position and its main geometrical features. The area occupied by the jet ﬂame was assessed by applying a temperature of 800 K, defining the jet ﬂame surface, based on observations of visible and infrared ﬂame images mentioned in \cite{palacios11}. Furthermore, it was generally found that, from the IR images, the ﬂame can be divided into three regions according to temperature behavior \cite{Vahid21} \cite{Gomez09}, which can be observed in Figure \ref{fig:horizontal}.

\begin{figure}[!htbp]
\subfloat[]{\includegraphics[width = 3.2in]{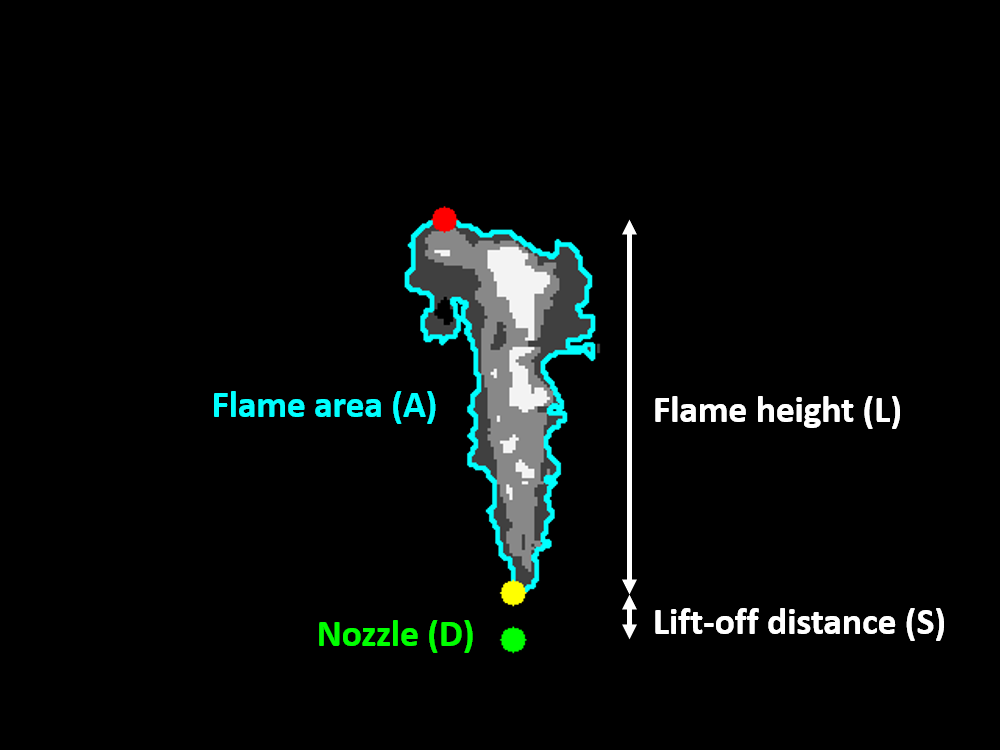}}
\label{fig:measureA}
\subfloat[]{\includegraphics[width = 3.2in]{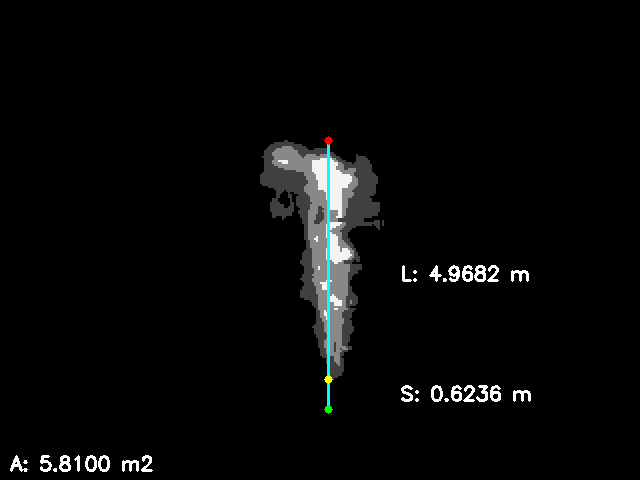}}
\label{fig:measureB}
\caption{Sample geometric characteristics extraction. Image (a) shows how the information is extracted from the segmentation produced by the UNet model. The contour from which the flame area ($A$) is obtained can be observed in blue. The highest point of the flame is represented as a red dot and indicating the tip of the flame. The base of the flame is represented as a yellow dot. The nozzle ($D$) is represented as a green dot at the lowest point. From those points the flame height ($L$) and the lift-off distance ($S$) are obtained. Image (b) is an example of this information extraction.}
\label{fig:measure}
\end{figure}

The IR videos were analyzed frame by frame (i.e. 4 frames per second), and for each IR frame a segmentation was performed to get: (i) the distance between the base of the stable ﬂame and the tip of the ﬂame, defined as the flame height ($L$); (ii) the distance between the pipe outlet diameter and the base of the stable ﬂame, defined as the lift-off distance ($S$); and (iii) the flame area ($A$). Obtaining these geometrical features is relevant to determine the probability of the two hazards associated with this type of fire accident: (i) thermal radiation, which can be very high at short distances, and (ii) flame impingement on a person or nearby equipment. The described geometrical features can be observed in Figure \ref{fig:measure}.

\subsection{Deep learning-based experimental set-up}
The segmentation of the jet fires and their geometrical information extraction was performed through a set of experiments. Initially, certain image processing methods were applied so that the images can then be used in machine learning algorithms and deep learning architectures. The resulting segmentation of each approach was then measured using certain metrics that evaluated the results. This section describes each one of these aspects.

\subsubsection{Data pre-processing }
The infrared (IR) recordings were extracted in Matlab ﬁles format. These ﬁles contain a temperature matrix corresponding to the IR image temperature values. The sets of image frames were exported from the IR video to test various segmentation models. To reduce the variance of the IR frame images, and to enhance their inherent characteristics, an image processing method was employed, known as image normalization. To achieve this, the mean and standard deviation per color channel has to be calculated over all images, then for each image, the channel values are divided by the maximum value of 255. From that result the obtained mean is subtracted and, finally, the values are divided by the standard deviation. This process leads to a quicker convergence of the deep learning models. The ground truth images were also changed into labeled images, by using a label id for each desired segmentation class. The RGB values are then changed to single-channel values by using the label ids as representation.

\begin{figure}[!htbp]
\centering
\subfloat[]{\includegraphics[width = 1.9in]{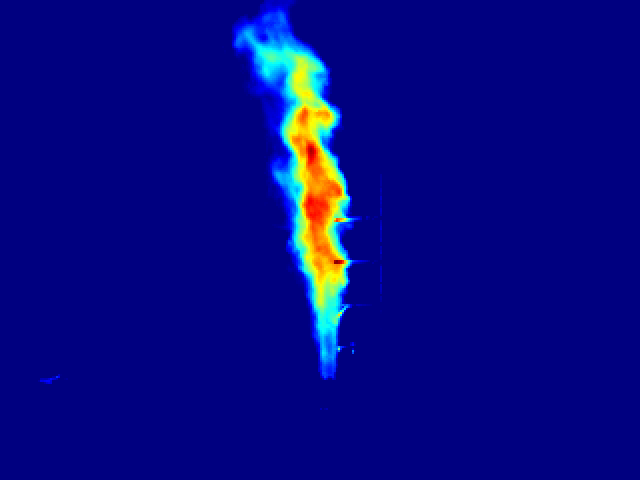}}
\qquad
\subfloat[]{\includegraphics[width = 1.9in]{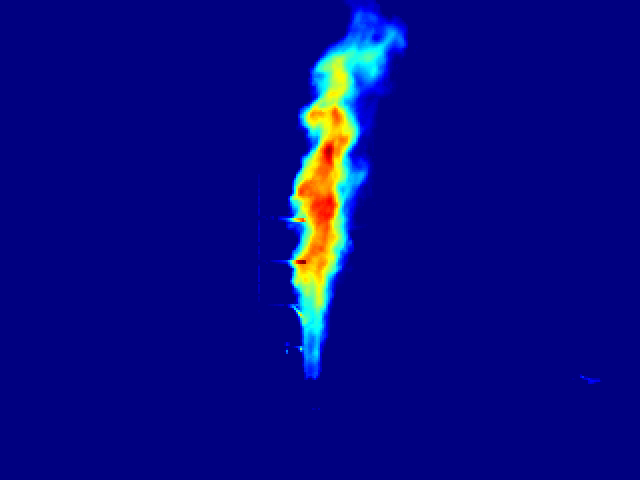}}
\qquad
\subfloat[]{\includegraphics[width = 1.43in]{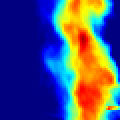}}
\caption{Examples of data augmentation. Figure (a) is the original infrared image. Figure (b) is the original image with horizontal flip. Figure (c) is a random crop of the original image.}
\label{fig:augmentation}
\end{figure}

Given that the training data set contains a small number of samples, data augmentation techniques were heavily applied to increase the models' performance and avoid over-fitting to the training data. This introduces variability to the input so that the model can perform well for instances not present in the training data. Three different sets of augmentation techniques were applied randomly throughout the experimentation workflow. These techniques are horizontal flipping, random scaling, and random cropping, some examples can be visualized in Figure \ref{fig:augmentation}.

\subsubsection{Evaluation metrics}

Based on the metric groups described in Taha and Hanbury \cite{taha15}, nine metrics were selected and separated into groups that describe their evaluation method, these metrics are enlisted in Table \ref{tab:metrics}. A Pearson correlation analysis was performed between the values of the selected metrics, calculated from a small number of segmentation examples, and a perceived ranking done by two experts in the context of the problem. This metric selection can be explored in more detail in a related paper \cite{perez21}.

\begin{table}[!htbp]
\centering
\caption{\label{tab:metrics} Summary of the metrics analysed through a selection process. The "Method" column describes the evaluation method group that the metric belongs to. Based on \cite{taha15}.}
\begin{tabular}{ll}
\centering
\textbf{Metric}            & \textbf{Method}              \\ \hline
Jaccard Index              & Spatial Overlap Based       \\ \hline
F-measure                  & Spatial Overlap Based      \\ \hline
Adjusted Rand Index      & Pair Counting Based         \\ \hline
Mutual Information         & Information Theoretic Based \\ \hline
Cohen's Kappa              & Probabilistic Based         \\ \hline
Hausdorff Distance         & Spatial Distance Based      \\ \hline
Mean Absolute Error        & Performance Based           \\ \hline
Mean Square Error          & Performance Based           \\ \hline
Peak Signal to Noise Ratio & Performance Based           \\ \hline
\end{tabular}
\end{table}

The metric with the highest correlation to the experts’ ranking was the Hausdorff Distance ($HD$), which is a dissimilarity measure mostly used when boundary delineation of the segmentation is important \cite{taha15}. The Hausdorff distance metric measures the extent to which each point of a segmentation lies near some point of the ground truth and vice versa; so it can be used to determine the degree of resemblance between the two images when superimposed on each other \cite{huttenlocher93}. A small resulting value is favored and preferable to maximize this metric. The Hausdorff Distance between two finite point sets $A$ and $B$ is defined as:

\begin{equation}
\label{eqn:haus}
HD(A,B)=max(h(A,B),h(B,A)).
\end{equation}
Where $h(A,B)$ is the directed Hausdorff distance \cite{taha15} given by:
\begin{equation}
h(A,B)=max_{a\epsilon A}min_{b\epsilon B}\left \| a-b \right \|.
\end{equation}

The Jaccard Index, also known as Intersection Over Union (IoU), has been also included in the analysis since it is the most commonly used metric for the evaluation of segmentation tasks. It calculates the area of overlap between the predicted segmentation and the ground truth divided by their area of union. For a multi-class problem, the mean IoU of the segmentation is calculated by averaging the IoU of each class \cite{rahman16}. In general, the closer the Jaccard Index is to a value of 1, the better the performance is. It is defined for the segments $S_{p}$ and $S_{t}$ as:

\begin{equation}
\label{eqn:jaccard}
JAC = \frac{S_{p} \cap S_{t}}{S_{p} \cup S_{t}} = \frac{TP}{TP+FP+FN}.
\end{equation}

Where $TP$ are the True Positives, $FP$ are the False Positives, and $FN$ are the False Negatives. These two metrics were used to compare the performance of some traditional machine learning segmentation models against a selected few Deep Learning segmentation models described in Section \ref{architectures}.

To further analyze the result of the segmentation given by a selected Deep Learning model, the jet fire's geometrical information was extracted from the generated mask and compared against the numeric experimental data. This evaluation was done using two different error measures, Mean Absolute Percentage Error (MAPE), and Root Mean Square Percentage Error (RMSPE). 

The  Mean Absolute Percentage Error measures the percentage for the average relative differences between $N$ values from the ground truth $x$ and the values from the predicted segmentation $y$. It does not consider the direction of the error and is used to measure accuracy for continuous variables \cite{Shcherbakov13}. It is defined as:

\begin{equation}
\label{eqn:mape}
MAPE = \frac{1}{N}\sum_{i=1}^{N} \frac{ |x_{i}-y_{i}|}{x_{i}} * 100.
\end{equation}

The Root Mean Square Percentage Error denotes the percentage for the relative quadratic difference between $N$ values from the ground truth $x$ and the values from the predicted segmentation $y$. It gives a relatively high weight to large errors \cite{Shcherbakov13}. It is defined as:

\begin{equation}
\label{eqn:rmspe}
RMSPE = \sqrt[]{\frac{1}{N}\sum_{i=1}^{N}\left ( \frac{x_{i}-y_{i} }{y_{i}}\right )^{2}} * 100.
\end{equation}

These two metrics can be used together to diagnose the variation in the errors. The RMSPE is usually larger or equal to the MAPE. With a greater difference between them, a greater variance in the individual errors can be inferred \cite{Shcherbakov13}.

Finally, to evaluate the statistical significance of the results, hypothesis testing was used. Because the MAPE is compared between models, it is expected that the majority of errors are small, and as they increase, their frequency is expected to decrease, so the data would be skewed to the left. To confirm this, a Quantile–Quantile (Q-Q) plot was used to illustrate the deviations of the data from a normal distribution \cite{yohai04}. Given the assumption of skewness of the data, a non-parametric test must be used, furthermore, since the errors are collected from the same experiments for each model, the type of test must be paired. Therefore, the statistical test used to compare the MAPE between models was the Wilcoxon signed-rank test \cite{oyeka12}. The null hypothesis is that there is no statistically significant difference between models, while the alternative hypothesis is that there is a statistically significant difference between the models. The significance level, which represents the probability of rejecting the null hypothesis when it is true, is defined as 0.1 for these experiments, given the small amount of data and the similarity between the architectures of UNet, Attention UNet, and UNet++.

\subsubsection{Segmentation approaches}
\label{architectures}

Based on the literature review, four different traditional image segmentation methods have been selected for this research, due to their characteristics and previous applications, to serve as a comparison baseline for the deep learning models explored, these methods are Global Thresholding, K-means clustering, Chan-Vese segmentation, and Gaussian Mixture Model (GMM). The deep learning architectures implemented were DeepLabv3 \cite{chen18}, SegNet \cite{Badrinarayanan2016}, UNet \cite{Ronneberger2015}, Attention UNet \cite{oktay18}, and UNet++ \cite{zhou19}. Further information on this selection can also be found in a related paper \cite{perez21}.

\subsubsection{Training}
Three thresholds were set for Global Thresholding, one for each of the areas of interest. They were computed automatically according to the histogram of grayscale training images with a median filter applied to make the values more uniform. The thresholds are based on the temperature intensities and defined as 31 to 85 for the outer zone, 101 to 170 for the middle zone, and 171 to 255 for the central zone (see Figure \ref{fig:horizontal}). The number of clusters for the K-Means algorithms was set to 4, according to the three areas of interest and the background. The final value of epsilon, which represents the required accuracy, was defined as 0.2. Four components were used for the Gaussian Mixture Model, according to the three areas of interest and the background. The covariance type was set to ‘tied’ since all components share the same general covariance matrix.

The parameters of the Chan-Vese algorithm were fine-tuned experimentally for each region. The parameter $\mu$ is the edge length weight parameter. Higher values of $\mu$ produce a rounder edge, while values closer to zero can detect smaller objects. The parameter $\lambda 1$ is the difference from the average weight parameter for the output region within the contour, while $\lambda 2$ is a similar parameter, for the output region outside the contour. When the $\lambda 1$ value is lower than the $\lambda 2$ value, the region within the contour will have a larger range of values, which is used when the background is very different from the segmented object in terms of distribution. Finally, $Tolerance$ represents the level set variation tolerance between iterations, used to detect if a solution was reached \cite{pascal12}. Table \ref{tab:parameters} shows the selected parameters.

\begin{table}[!htbp]
\caption{Parameter selection for the Chan-Vese segmentation algorithm.}
\label{tab:parameters}
\centering
\begin{tabular}{cccc}
\hline
\textbf{Parameter}   & \textbf{Outer Zone} & \textbf{Middle Zone} & \textbf{Central Zone} \\ \hline
\textbf{\textmu}      & 0.3         & 0.01      & 0.02      \\ \hline
\textbf{\textlambda 1} & 1.0        & 0.5       & 0.5       \\ \hline
\textbf{\textlambda 2} & 1.5        & 2.0       & 2.5       \\ \hline
\textbf{$Tolerance$}   & 0.001        & 0.002     & 0.0009    \\ \hline
\end{tabular}
\end{table}

The training of the Deep Learning models was done in an Nvidia DGX-1 Station system that has 8 NVIDIA Volta-based GPUs. The models were trained using Early Stopping for up to 5000 epochs with a learning rate and weight decay of 0.0001. An ADAM optimizer is used for the learning rate and a batch size of 4 is used for training. The loss function employed is a weighted cross-entropy loss function that combines Log Softmax and Negative Log Likelihood (NLL) loss. The initialization of the weights was calculated based on the ENet custom class weighting scheme \cite{paszke16}. The values were defined as 1.59 for the background, 10.61 for the outer zone, 17.13 for the middle zone, and 22.25 for the central zone. The data set of 201 images is split in 80\% for training and 20\% for validation and testing, resulting in a total of 161 images for training, 20 images for validation, and 20 images for testing.

\section{Results and Discussion}
\label{sec:results}

\begin{figure}[!htbp]
\centering
\includegraphics[width=0.9\textwidth]{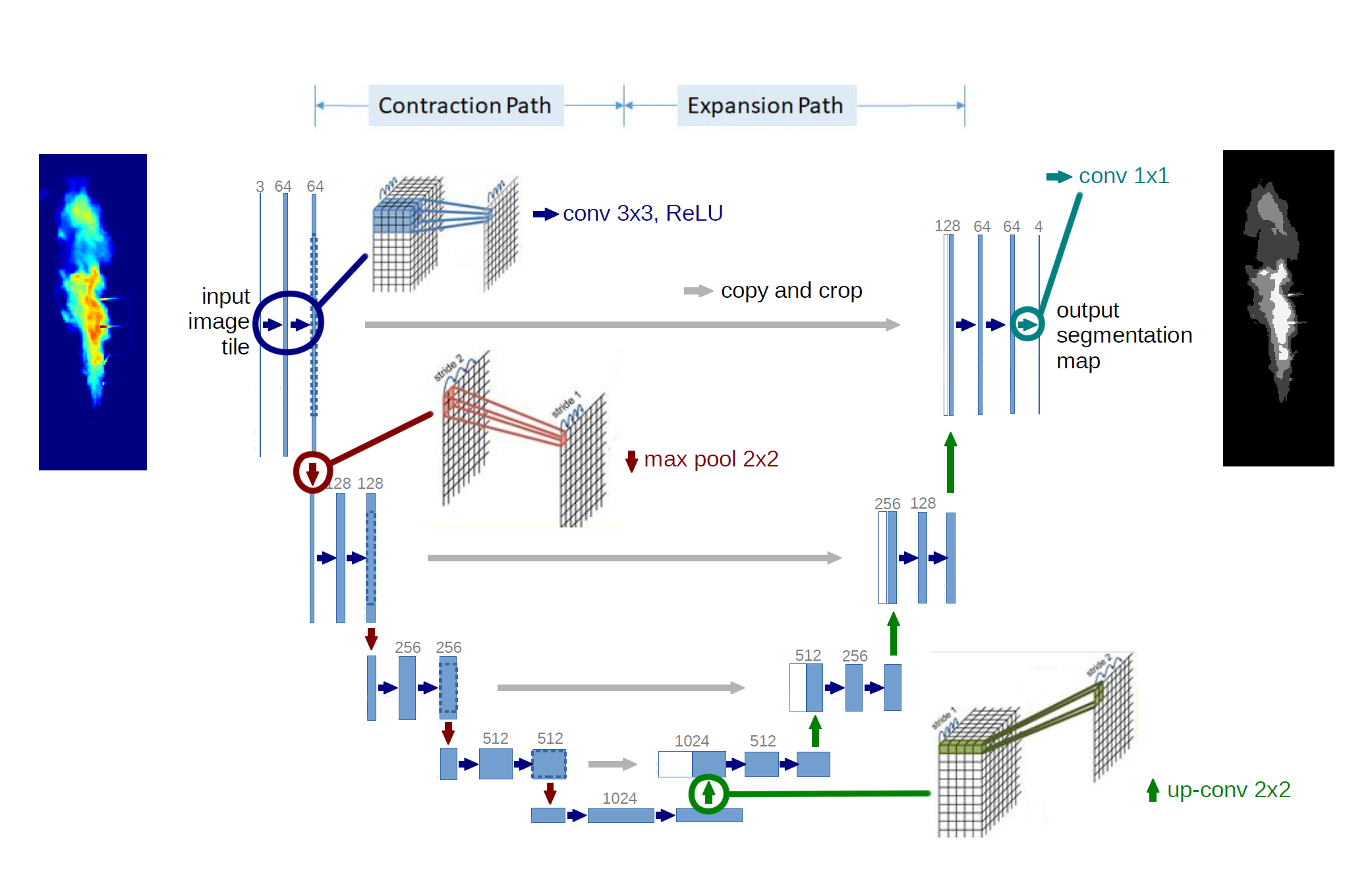}
\caption{UNet architecture. Each blue box corresponds to a multi-channel feature map. The number of channels is denoted on top of the box. White boxes represent copied feature maps. The gray arrows denote the different operations \cite{Ronneberger2015}.}
\label{fig:UNet}
\end{figure}

Out of all the approaches discussed in Section \ref{architectures}, the Deep Learning architectures generally outperformed the traditional segmentation methods. Deeplabv3, SegNet, and UNet are substantially encoder-decoder architectures. The encoder uses downsampling to reduce the spatial resolution of the input and generate lower resolution feature mappings; this makes them computationally efficient and improves the learning of relevant features. The decoder upsamples those feature representations, which generates a full resolution segmentation map that offers pixel-wise predictions \cite{xing20}. The spatial information recreated during upsampling can be imprecise, so UNet uses skip connections that combine spatial information between its contracting and expanding path \cite{Ronneberger2015}, which provides a more precise localization of the regions, compared to the other architectures. This architecture can be visualized in Figure \ref{fig:UNet}.

Attention UNet improves on the UNet architecture with the use of additive soft attention. The initial layers of UNet generate poor feature representations and that causes skip connections to bring across several redundant low-level feature maps. The soft attention implemented in the skip connections of Attention UNet actively suppresses activations in irrelevant regions, reducing the number of redundant features brought across \cite{oktay18}. UNet++ also improves on the skip connections used in U-Net, since the direct combination of high-resolution feature maps from the encoder to the decoder can lead to the concatenation of semantically different feature maps. UNet++  is based on nested and dense skip connections that generally try to bridge the semantic gap between the feature maps before concatenation, effectively capturing fine-grained details \cite{zhou19}.

\begin{figure}[!htbp]
\subfloat[]{\includegraphics[width = 3in]{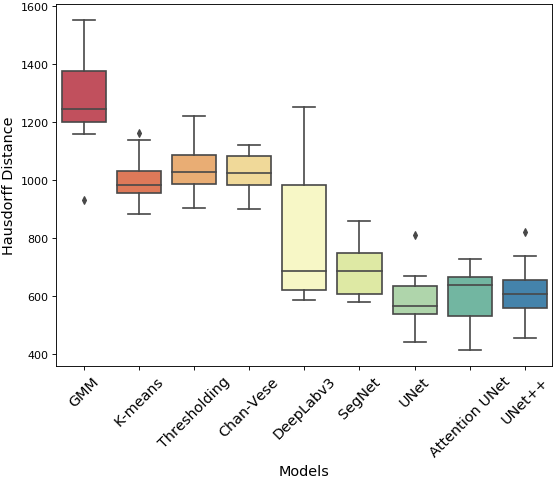}}
\label{fig:haus}
\subfloat[]{\includegraphics[width = 3in]{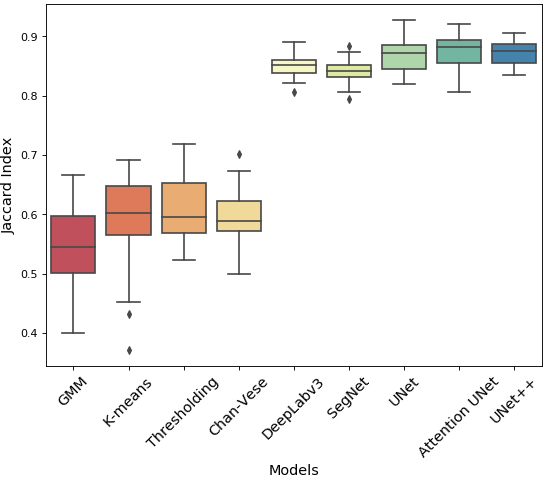}}
\label{fig:jacc}
\caption{Box and whisker plots for the metrics of each segmentation. The values are from the result of the models mentioned in Section \ref{architectures}. Figure (a) shows the Hausdorff Distance results. Figure (b) shows the Jaccard Index results.}
\label{fig:comparison}
\end{figure}

\begin{figure}[!htbp]
\centering
\subfloat[]{\includegraphics[width = 1.9in]{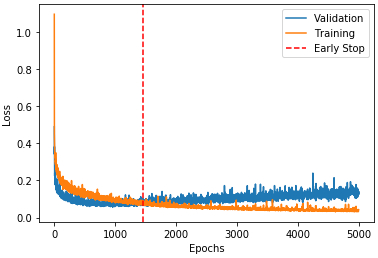}}
\qquad
\subfloat[]{\includegraphics[width = 1.9in]{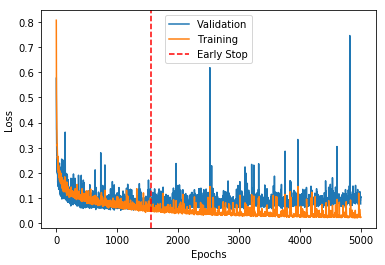}}
\qquad
\subfloat[]{\includegraphics[width = 1.9in]{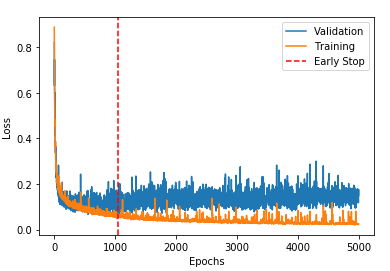}}
\caption{Loss Function values of UNet, Attention UNet, and UNet++ models. Training loss is presented as a thin orange curve, and validation loss is presented as a thick blue curve. The moment early stopping takes place is denoted as a dashed vertical red line. Figure (a) illustrates the UNet loss function values. Early stopping took place at 1460 epochs. Figure (b) illustrates the Attention UNet loss function values. Early stopping took place at 1560 epochs. Figure (c) illustrates the UNet++ loss function values. Early stopping took place at 1055 epochs.}
\label{fig:loss}
\end{figure}

\begin{figure}[!htbp]
\centering
\subfloat[]{\includegraphics[width = 1.3in]{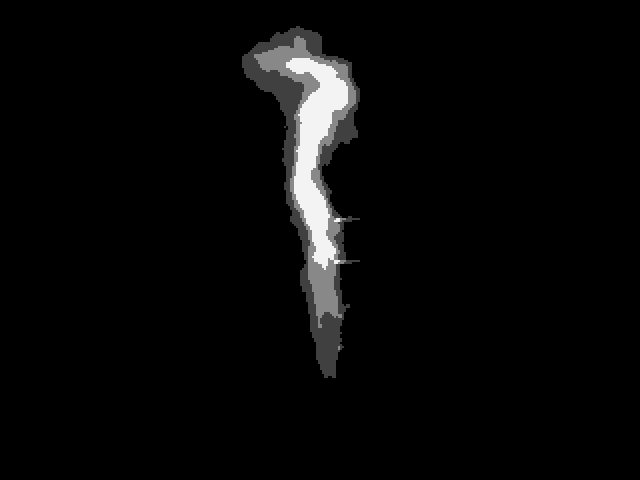}}
\qquad
\subfloat[]{\includegraphics[width = 1.3in]{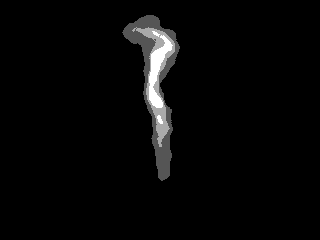}} \\

\subfloat[]{\includegraphics[width = 1.3in]{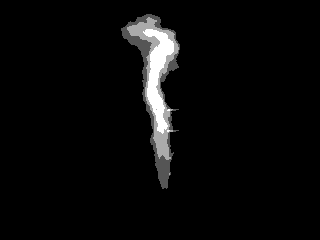}}
\qquad
\subfloat[]{\includegraphics[width = 1.3in]{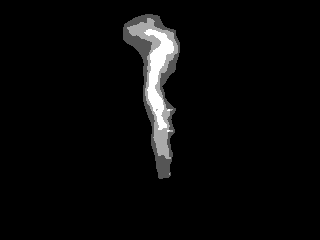}} \\

\subfloat[]{\includegraphics[width = 1.3in]{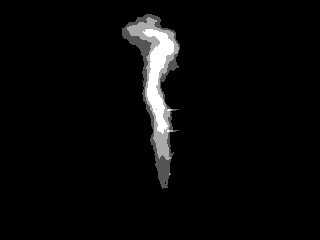}}
\qquad
\subfloat[]{\includegraphics[width = 1.3in]{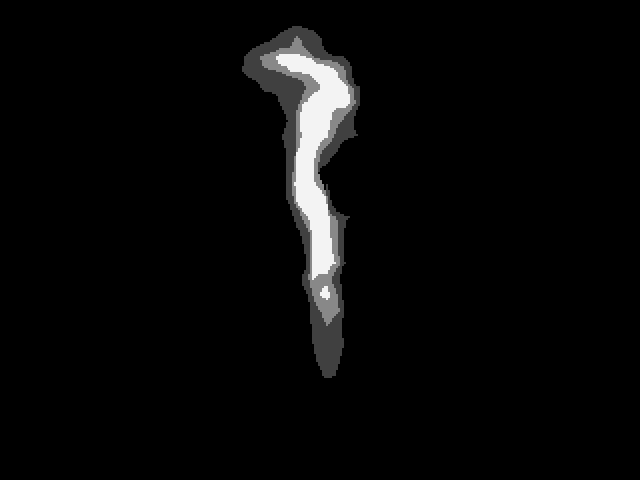}} \\

\subfloat[]{\includegraphics[width = 1.3in]{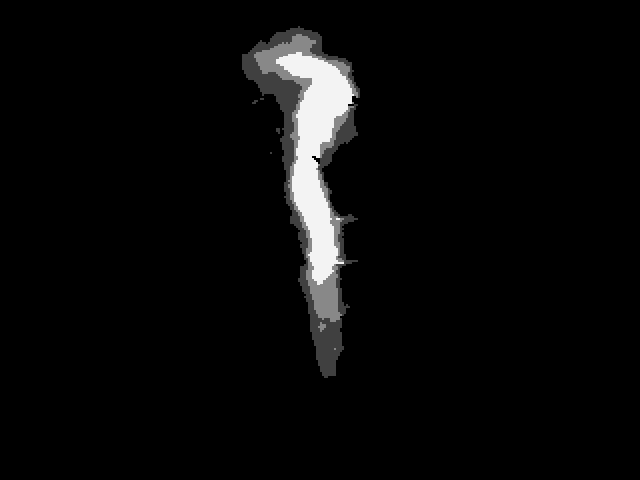}}
\qquad
\subfloat[]{\includegraphics[width = 1.3in]{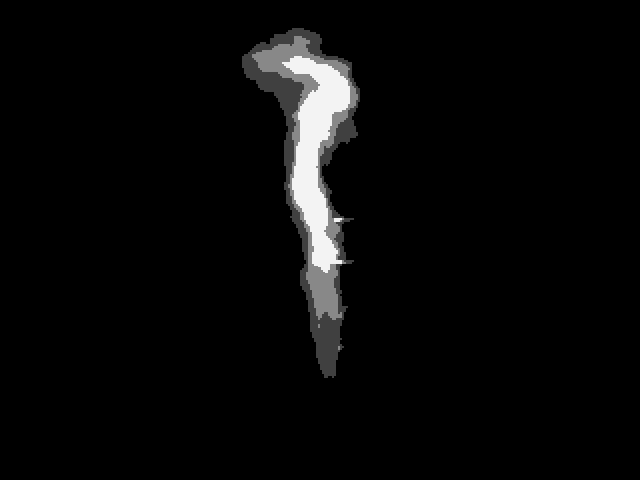}} \\

\subfloat[]{\includegraphics[width = 1.3in]{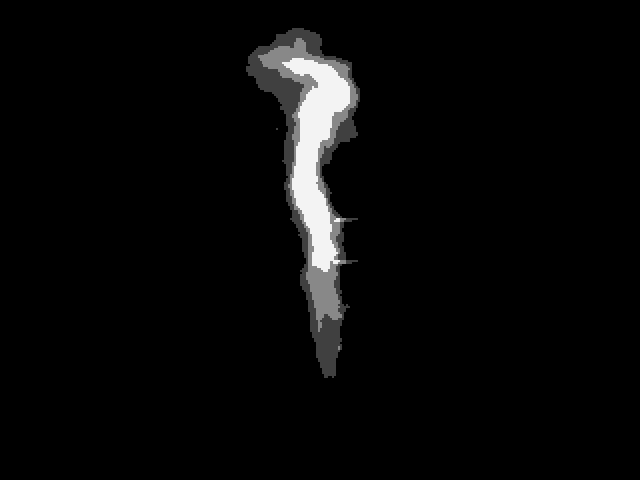}}
\qquad
\subfloat[]{\includegraphics[width = 1.3in]{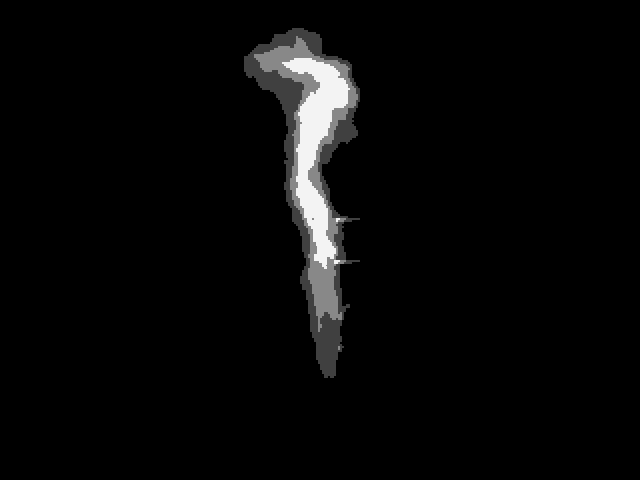}} \\

\caption{Sample segmentation masks. The segmentation results are obtained from the different models described in Section \ref{architectures}. Figure (a) is the Ground Truth segmentation. Figure (b) is the Global Thresholding segmentation. Figure (c) is the K-means segmentation. Figure (d) is the Chan-Vese segmentation. Figure (e) is the Gaussian Mixture Model segmentation. Figure (f) is the DeepLabv3 segmentation. Figure (g) is the SegNet segmentation. Figure (h) is the UNet segmentation. Figure (i) is the Attention UNet segmentation. Figure (j) is the UNet++ segmentation. Slight differences can be observed along the central and middle zones, especially at the top and bottom parts of the flame.}
\label{fig:segmentations}
\end{figure}

UNet obtained the lowest mean Hausdorff Distance value of 586.46. Given the performance observed, two recent improvements to the UNet architecture were further explored, Attention UNet and UNet++, mentioned as well in Section \ref{architectures}.  Attention UNet obtained the highest mean Jaccard Index, with a value of 0.8746. UNet++ achieved the least time to segment all 201 images from the dataset, using only 14.6 seconds. These results can be observed in Table \ref{tab:model-results} and Figure \ref{fig:comparison}. Early stopping took place at 1460 epochs for UNet, at 1560 epochs for Attention UNet, and at 1055 for UNet++. The loss function values of these models can be observed in Figure \ref{fig:loss}. The difference in segmentation between all the models is illustrated as an example in Figure \ref{fig:segmentations}.

\begin{table}[!htbp]
\caption{The mean Hausdorff Distance and Jaccard Index for all the methods mentioned in Section \ref{architectures} across the testing set of 20 images. The time it took for each method to segment all 201 images from the whole data set is included in the Time column, it is expressed in seconds. The best results obtained are in bold.}
\label{tab:model-results}
\centering
\begin{tabular}{cccc}
\hline
\textbf{Method}       & \textbf{Hausdorff Distance} & \textbf{Jaccard Index} & \textbf{Time (s)} \\ \hline
GMM          & 1288.10                     & 0.5460                 & 2723.8           \\ \hline
K-means      & 1000.63                     & 0.5876                 & 3035.1           \\ \hline
Thresholding & 1029.08                     & 0.6092                 & 30.7             \\ \hline
Chan-Vese    & 1031.90                     & 0.5906                 & 18177.5          \\ \hline
Deeplabv3      & 784.86                    & 0.8502                 & 17.1             \\ \hline
Segnet       & 692.73                      & 0.8405                 & 16.4             \\ \hline
UNet         & \textbf{586.46}             & 0.8681                 & 15.7             \\ \hline
Attention UNet         & 601.05                      & \textbf{0.8746}                 & 17.7             \\ \hline
UNet++         & 615.63                      & 0.8722                 & \textbf{14.6}             \\ \hline
\end{tabular}
\end{table}

Based on these results, 300 infrared images of vertical propane jet fires were segmented by using the UNet, Attention UNet, and UNet++ models. From that segmentation, the jet flame height, $L$, the flame area, $A$, and the lift-toff distance, $S$, have been obtained as shown in the Figure \ref{fig:measure} of Section \ref{sec:boundary}.

\subsection{Flame height}
\begin{figure}[!htbp]
\subfloat[]{\includegraphics[width = 3in]{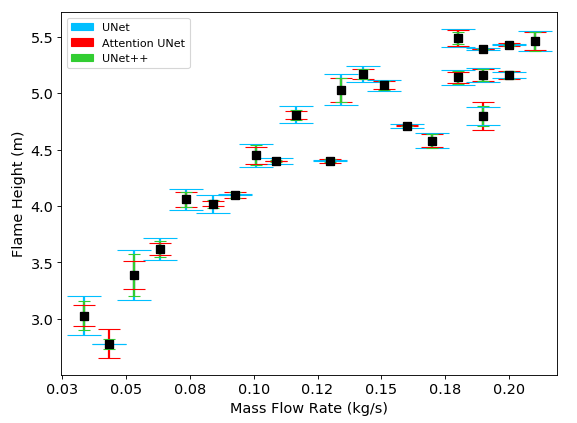}} 
\subfloat[]{\includegraphics[width = 3.05in]{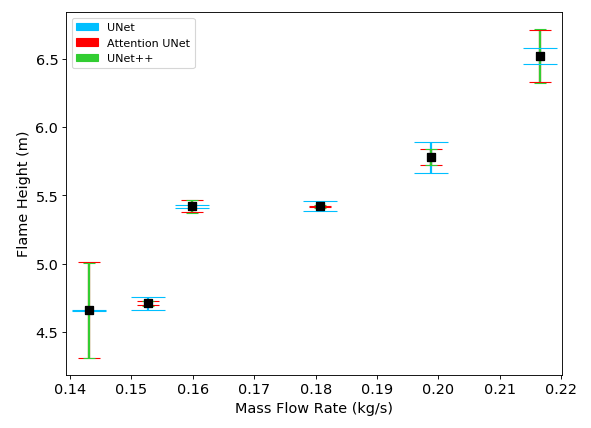}}\\
\subfloat[]{\includegraphics[width = 3in]{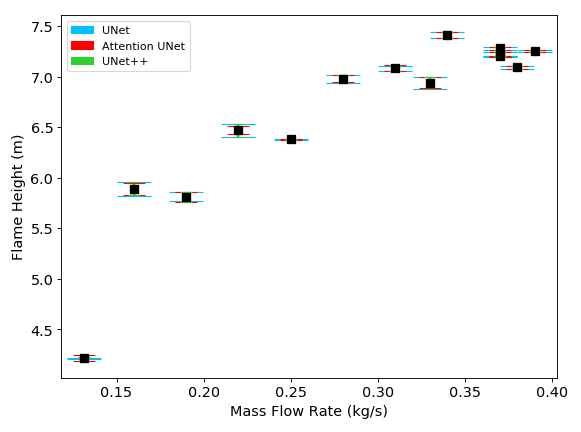}}
\subfloat[]{\includegraphics[width = 3.1in]{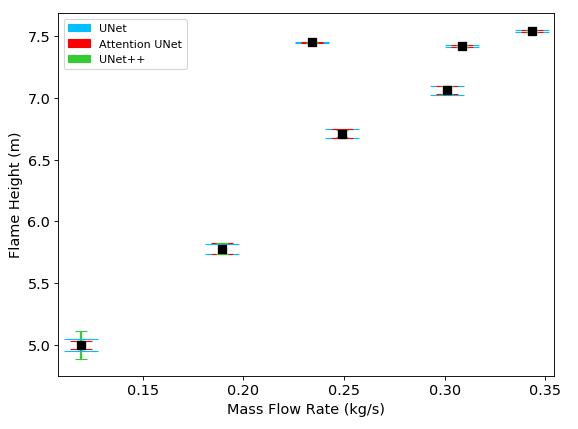}} 
\caption{Detected flame height for all the experiments analysed. The error bars in blue correspond to the results from UNet, the ones in red are from Attention UNet, and the ones in green are from UNet++. Figure (a) shows the results for the experiment with pipe outlet diameter of 0.01275 m. Figure (b) shows the results for the experiment with pipe outlet diameter of 0.015 m. Figure (c) shows the results for the experiment with pipe outlet diameter of 0.02 m. Figure (d) shows the results for the experiment with pipe outlet diameter of 0.03 m.}
\label{fig:H}
\end{figure}

The jet flame height, $L$, was measured from the base of the flame to the tip of the flame. In Figure \ref{fig:H}, the flame height was plotted against the mass flow rate for all the present subsonic and sonic data, vertically released into still air from four circular nozzles (i.e. 0.01275, 0.015, 0.02, and 0.03 m).

From Figure \ref{fig:H} it can be observed that the experimental vertical jet flames go up to 7.7 m in height, and show a good agreement with the obtained predictions. In general, the flame height tends to increase with the mass flow rate; however, there is some variance that reflects the turbulence of propane flames. There are fewer errors visible as the diameter of the nozzle gets larger. Table \ref{tab:height} shows the errors of the present experimental jet flame heights, obtained from the segmentation of the UNet, Attention UNet, and UNet++ models. The error is presented for each analyzed experiment and identified by the pipe outlet diameter. The presented errors are the Mean Absolute Percentage Error (MAPE) and the Root Mean Square Percentage Error (RMSPE). 

\begin{table}[!htbp]
\centering
\caption{The error of the flame height obtained from different segmentation models and for each experiment analysed.}
\label{tab:height}
\begin{tabular}{lllll}
\hline
\multicolumn{5}{c}{\textbf{Height}}                                                                                                                \\ \hline
\textbf{\begin{tabular}[c]{@{}l@{}}Pipe outlet \\ diameter (m)\end{tabular}} & \textbf{0.01275} & \textbf{0.015} & \textbf{0.02}  & \textbf{0.03}  \\ \hline
\multicolumn{5}{c}{\textbf{UNet}}                                                                                                                  \\ \hline
MAPE                                                                         & 6.8\%            & \textbf{4.5\%} & 2.9\%          & 2.5\%          \\ \hline
RMSPE                                                                        & 8.7\%            & \textbf{5.8\%} & 3.7\%          & 3.1\%          \\ \hline
\multicolumn{5}{c}{\textbf{Attention UNet}}                                                                                                        \\ \hline
MAPE                                                                         & \textbf{5.5\%}   & 11.1\%           & \textbf{2.9\%} & \textbf{2.3\%} \\ \hline
RMSPE                                                                        & \textbf{6.7\%}   & 16.7\%         & \textbf{3.4\%} & \textbf{2.7\%} \\ \hline
\multicolumn{5}{c}{\textbf{UNet++}}                                                                                                                \\ \hline
MAPE                                                                         & 5.6\%            & 11.4\%         & 2.9\%          & 3.3\%          \\ \hline
RMSPE                                                                        & 7.0\%            & 16.7\%         & 3.6\%          & 4.9\%          \\ \hline
\end{tabular}
\end{table}

For the 0.01275 m experiments, Attention UNet achieved the lowest errors and error variations, with a MAPE of 5.5\% and RMSPE of 6.7\%. UNet++ had a similar MAPE, but showed a higher RMSPE, denoting a higher variance in the errors.

For the 0.015 m experiments, UNet achieved the lowest errors and error variations with a MAPE of 4.5\% and RMSPE of 5.8\%. Attention UNet and UNet++ obtained larger errors in comparison, which mostly happened with the lowest and highest mass flow rates.

For the 0.02 m experiments, Attention UNet achieved the lowest errors and error variations with a MAPE of 2.9\% and RMSPE of 3.4\%. UNet and UNet++ obtained the same MAPE, but their RMSPE was higher in comparison, which means that the variation of the errors obtained from those models is higher.

For the 0.03 m experiments, Attention UNet achieved the lowest errors and error variations with a MAPE of 2.3\% and RMSPE of 2.7\%. UNet obtained a similar MAPE but higher RMSPE, and UNet++ obtained higher overall errors.

\begin{figure}[!htbp]
\centering
\subfloat[]{\includegraphics[width = 2.5in]{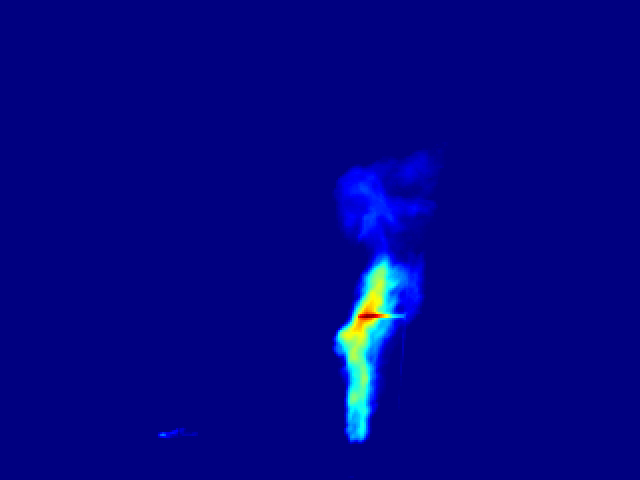}}
\qquad
\subfloat[]{\includegraphics[width = 2.5in]{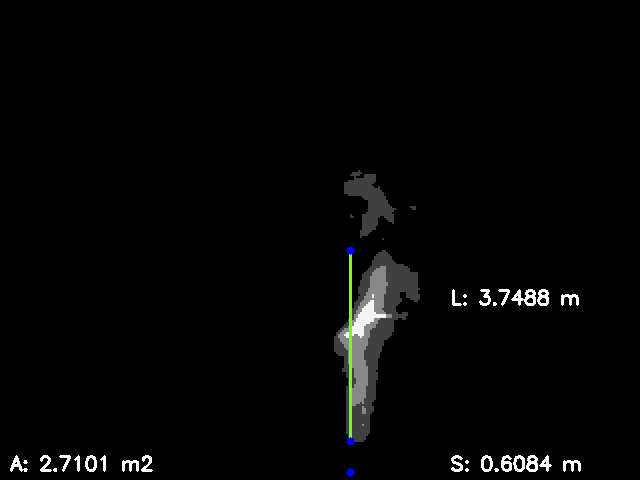}} \\
\subfloat[]{\includegraphics[width = 2.5in]{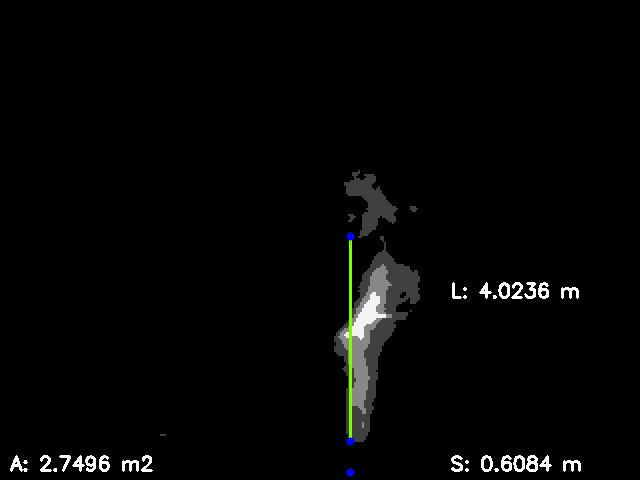}}
\qquad
\subfloat[]{\includegraphics[width = 2.5in]{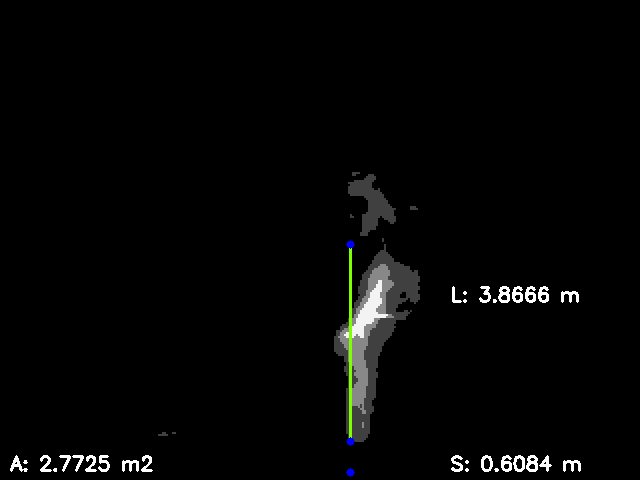}}
\caption{Example segmentation of a jet flame with outlet pipe diameter of 0.015 m and mass flow rate of 0.143 kg/s. The extracted geometric characteristics of flame area ($A$), flame height ($L$), and lift-off distance ($S$) are included. Figure (a) shows the original infrared image. Figure (b) shows the UNet segmentation results. Figure (c) shows the Attention UNet segmentation results. Figure (d) shows the UNet++ segmentation results.}
\label{fig:15mmexample}
\end{figure}

Attention UNet obtained the lowest MAPE and RMSPE for almost all pipe outlet diameter experiments, the only exception was the 0.015 m case. For that experiment, UNet was the best model by a high margin. This large difference mostly happens during the lowest and highest mass flow rates, where the jet fires show sections that are detached from the main flame and add uncertainty to its limits. Attention UNet and UNet++ over-segment these parts and lead to such high errors. An example of this can be observed in Fig. \ref{fig:15mmexample}.

\subsection{Flame area}

\begin{figure}[!htbp]
\subfloat[]{\includegraphics[width = 3in]{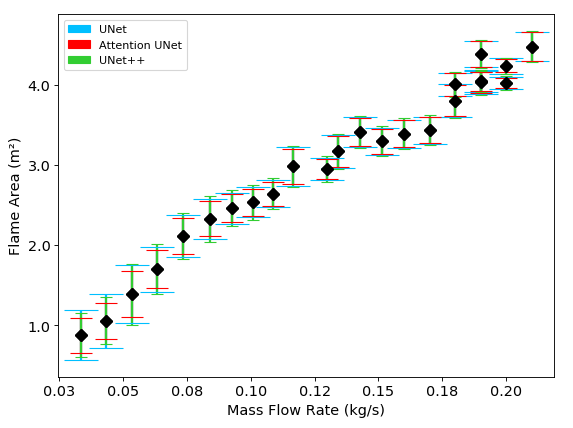}} 
\subfloat[]{\includegraphics[width = 3.04in]{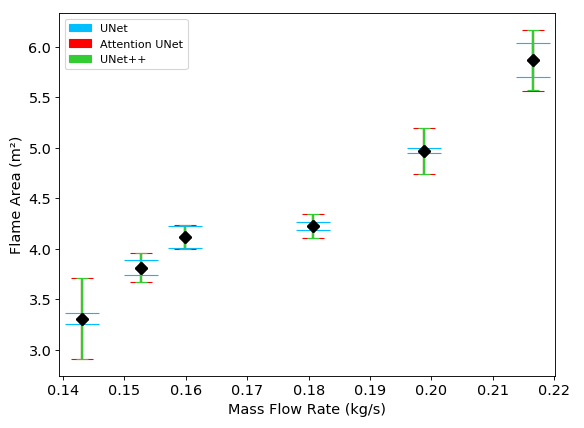}}\\
\subfloat[]{\includegraphics[width = 3in]{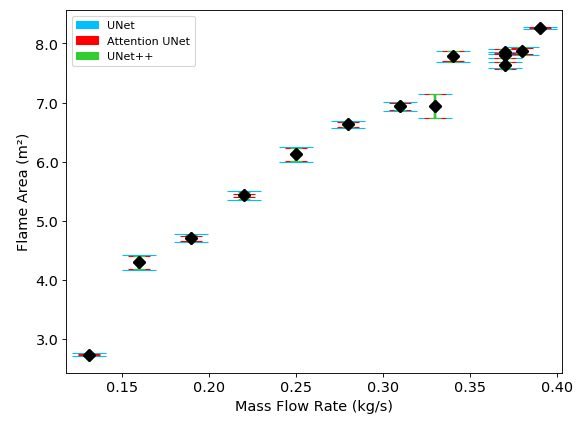}}
\subfloat[]{\includegraphics[width = 3in]{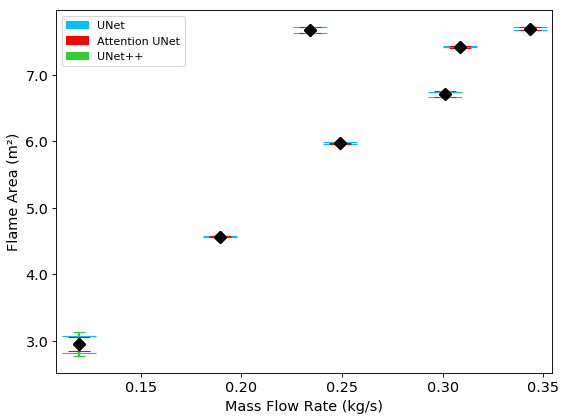}} 
\caption{Detected flame area for all the experiments analysed. The error bars in blue correspond to the results from UNet, the ones in red are from Attention UNet, and the ones in green are from UNet++. Figure (a) shows the results for the experiment with pipe outlet diameter of 0.01275 m. Figure (b) shows the results for the experiment with pipe outlet diameter of 0.015 m. Figure (c) shows the results for the experiment with pipe outlet diameter of 0.02 m. Figure (d) shows the results for the experiment with pipe outlet diameter of 0.03 m.}
\label{fig:A}
\end{figure}

To calculate the radiation from a jet ﬁre on a certain target, the ﬂame area and size are required. Figure \ref{fig:A} shows the jet flame area, $A$, plotted against the mass flow rate for all the present experimental data, involving the four circular nozzles of 0.01275 m, 0.015 m, 0.02 m, and 0.03 m.

From Figure \ref{fig:A} it can be observed that the experimental vertical jet flames reach an area of up to 8.3 m$^2$, which is also approximated with decent accuracy by the obtained predictions. The most prominent errors are observed in the 0.01275 m experiment; however, they seem to be consistent, which could mean that there is a common problem that affects the general accuracy. This behavior could be explained by the turbulent nature of the flame at such a small outlet diameter. Table \ref{tab:area} shows the errors of the present experimental jet flame areas, obtained from the segmentation of the UNet, Attention UNet, and UNet++ models. The error is presented for each analyzed experiment and identified by the pipe outlet diameter. The presented errors are the Mean Absolute Percentage Error (MAPE) and the Root Mean Square Percentage Error (RMSPE).

\begin{table}[!htbp]
\centering
\caption{The error of the flame area obtained from different segmentation models and for each experiment analysed.}
\label{tab:area}
\begin{tabular}{lllll}
\hline
\multicolumn{5}{c}{\textbf{Area}}                                                                                                                  \\ \hline
\textbf{\begin{tabular}[c]{@{}l@{}}Pipe outlet \\ diameter (m)\end{tabular}} & \textbf{0.01275} & \textbf{0.015} & \textbf{0.02}  & \textbf{0.03}  \\ \hline
\multicolumn{5}{c}{\textbf{UNet}}                                                                                                                  \\ \hline
MAPE                                                                         & 19.9\%           & \textbf{7.8\%} & 7.9\%          & 3.9\%          \\ \hline
RMSPE                                                                        & 21.1\%           & \textbf{9.1\%} & 9.2\%          & 5.5\%          \\ \hline
\multicolumn{5}{c}{\textbf{Attention UNet}}                                                                                                        \\ \hline
MAPE                                                                         & \textbf{17.3\%}  & 21.9\%         & \textbf{6.5\%} & \textbf{3.6\%} \\ \hline
RMSPE                                                                        & \textbf{18.0\%}  & 24.3\%         & \textbf{8.2\%} & \textbf{4.8\%} \\ \hline
\multicolumn{5}{c}{\textbf{UNet++}}                                                                                                                \\ \hline
MAPE                                                                         & 21.2\%           & 21.9\%         & 7.1\%          & 4.6\%          \\ \hline
RMSPE                                                                        & 22.2\%           & 24.3\%         & 8.6\%          & 7.3\%          \\ \hline
\end{tabular}
\end{table}

For the 0.01275 m experiments, Attention UNet achieved the lowest errors and error variations, with a MAPE of 17.3\% and RMSPE of 18.0\%. As mentioned before, these are the largest errors obtained overall; however,  the small difference between the MAPE and RMSPE, shows that there is not a high variance between errors, alluding to a consistent problem derived from the flames in this specific case.

For the 0.015 m experiments, UNet achieved the lowest errors and error variations with a MAPE of 7.8\% and RMSPE of 9.1\%. Similar to what happened with the predicted height, Attention UNet and UNet++ obtained larger errors in comparison, which mostly happened with the lowest mass flow rate and the two highest mass flow rates.

For the 0.02 m experiments, Attention UNet achieved the lowest errors with a MAPE of 6.5\% and RMSPE of 8.2\%. UNet obtained the highest MAPE, but the lowest difference between MAPE and RMSPE, which means that it has smaller error variations.

For the 0.03 m experiments, Attention UNet achieved the lowest errors and error variations with a MAPE of 3.6\% and RMSPE of 4.8\%. UNet obtained a similar MAPE but higher RMSPE, and UNet++ obtained higher overall errors.

In a similar way to the flame height results, Attention UNet obtained the lowest MAPE and RMSPE for the area predictions of almost all pipe outlet diameter experiments, the only exception was the 0.015 m case. For that experiment, UNet was the best model by a high margin. The area was the geometric characteristic that obtained the highest errors for the experiments presented in this paper; however, this can be explained by how the experimental area is usually obtained from a minimum bounding box. By segmenting the flame, the areas without flame that may have been included in the minimum bounding box are excluded from the prediction. Even so, the highest error still allows for an approximate representation of the experimental area.

\subsection{Lift-off distance}

\begin{figure}[!htbp]
\subfloat[]{\includegraphics[width = 3in]{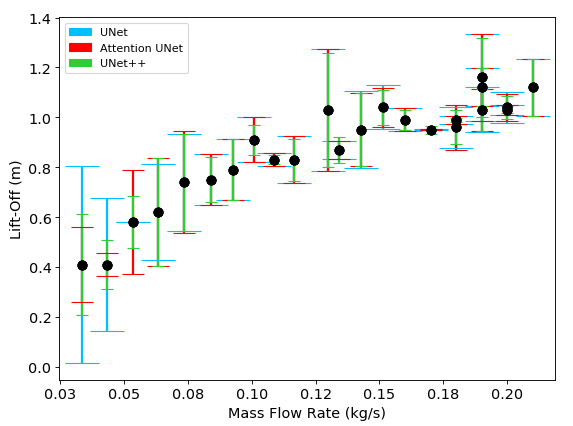}} 
\subfloat[]{\includegraphics[width = 3.04in]{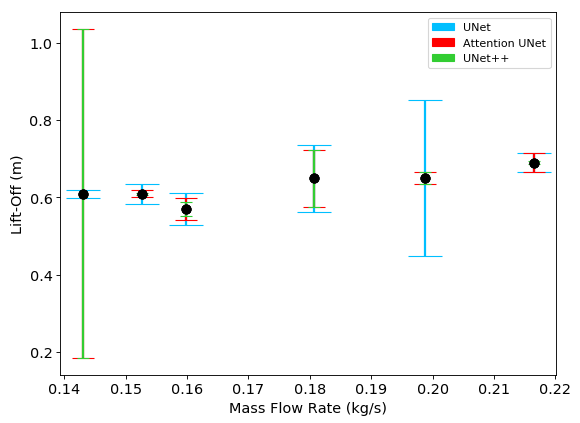}}\\
\subfloat[]{\includegraphics[width = 3in]{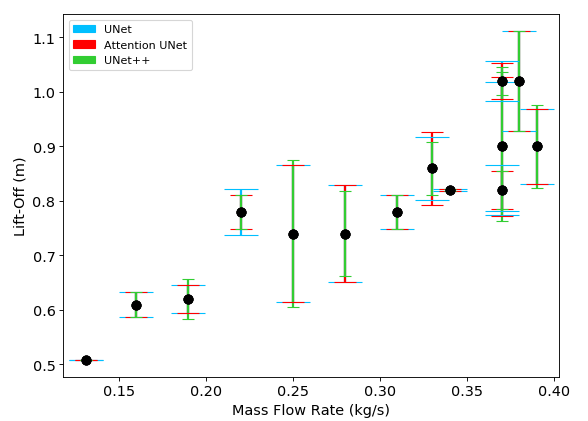}}
\subfloat[]{\includegraphics[width = 3.04in]{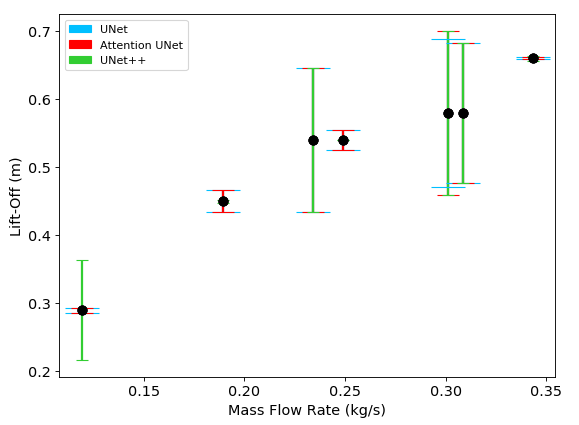}} 
\caption{Detected lift-off distance for all the experiments analysed. The error bars in blue correspond to the results from UNet, the ones in red are from Attention UNet, and the ones in green are from UNet++. Figure (a) shows the results for the experiment with pipe outlet diameter of 0.01275 m. Figure (b) shows the results for the experiment with pipe outlet diameter of 0.015 m. Figure (c) shows the results for the experiment with pipe outlet diameter of 0.02 m. Figure (d) shows the results for the experiment with pipe outlet diameter of 0.03 m.}
\label{fig:LO}
\end{figure}

The lift-off distance, $S$, was measured from the outlet orifice pipe, to the base of the lifted stabilized jet flame. In Figure \ref{fig:LO}, the lift-off distance was plotted against the mass flow rate for the present experimental data, involving the four circular nozzles of 0.01275 m, 0.015 m, 0.02 m, and 0.03 m.

From Figure \ref{fig:LO} it can be seen that the lift-off distances go up to 1.1 m, and also show good agreement between predicted and experimental values for the present relatively large turbulent jet flames released in sonic and subsonic regimes. It is important to note that lift-off distances have a high sensitivity to the mixing processes and unstable ﬂame ﬂuctuations, which is reflected in Figure \ref{fig:LO}.

Table \ref{tab:liftoff} shows the errors for the present experimental lift-off distances,   obtained from the segmentation of the UNet, Attention UNet, and UNet++ models. The error is presented for each analyzed experiment and identified by the pipe outlet diameter. The presented errors are the Mean Absolute Percentage Error (MAPE) and the Root Mean Square Percentage Error (RMSPE).

\begin{table}[!htbp]
\centering
\caption{The error of the lift-off distance obtained from different segmentation models and for each experiment analysed.}
\label{tab:liftoff}
\begin{tabular}{lllll}
\hline
\multicolumn{5}{c}{\textbf{Lift-Off}}                                                                                                              \\ \hline
\textbf{\begin{tabular}[c]{@{}l@{}}Pipe outlet \\ diameter (m)\end{tabular}} & \textbf{0.01275} & \textbf{0.015} & \textbf{0.02}  & \textbf{0.03}  \\ \hline
\multicolumn{5}{c}{\textbf{UNet}}                                                                                                                  \\ \hline
MAPE                                                                         & 11.2\%           & \textbf{6.5\%} & \textbf{5.4\%} & \textbf{5.0\%} \\ \hline
RMSPE                                                                        & 14.6\%           & \textbf{9.2\%} & \textbf{6.6\%} & \textbf{7.0\%} \\ \hline
\multicolumn{5}{c}{\textbf{Attention UNet}}                                                                                                        \\ \hline
MAPE                                                                         & 10.2\%           & 9.6\%          & 5.4\%          & 5.2\%          \\ \hline
RMSPE                                                                        & 12.2\%           & 17.8\%         & 6.7\%          & 7.2\%          \\ \hline
\multicolumn{5}{c}{\textbf{UNet++}}                                                                                                                \\ \hline
MAPE                                                                         & \textbf{9.6\%}   & 9.0\%          & 5.4\%          & 5.8\%          \\ \hline
RMSPE                                                                        & \textbf{11.6\%}  & 17.7\%         & 6.8\%          & 7.7\%          \\ \hline
\end{tabular}
\end{table}

For the 0.01275 m experiments, UNet++ achieved the lowest errors and error variations, with a MAPE of 9.6\% and RMSPE of 11.6\%. This is the only instance where UNet++ has a better performance than the other two models. Attention UNet has a slightly higher MAPE, but it obtains the same difference between MAPE and RMSPE as the UNet++ results, which means that they have similar error variations.

For the 0.015 m experiments, UNet achieved the lowest errors and error variations with a MAPE of 6.5\% and RMSPE of 9.2\%. Similar to what happened with the predicted height and area, Attention UNet and UNet++ obtained larger errors in comparison, which mostly happened with the lowest mass flow rate.

For the 0.02 m experiments, UNet achieved the lowest errors with a MAPE of 5.4\% and RMSPE of 6.6\%. Attention UNet and UNet++ obtained the same MAPE, but slightly higher RMSPE, which means that they present a higher error variation.

For the 0.03 m experiments, UNet achieved the lowest errors and error variations with a MAPE of 5.0\% and RMSPE of 7.0\%. Attention UNet and UNet++ obtained slightly higher MAPE but relatively homogeneous differences between MAPE and RMSPE, which means that they all presented similar error variances.

The lift-off distance presented a large variance between errors, which can be explained by the turbulence of these open field flames. Contrary to the flame height and area results, UNet obtained the lowest MAPE and RMSPE for the lift-off distance predictions of almost all pipe outlet diameter experiments, the only exception was the 0.01275 m case; for that experiment, UNet++ was the best model.

\subsection{Statistical test}

\begin{figure}[!htbp]
\centering
\subfloat[]{\includegraphics[width = 1.9in]{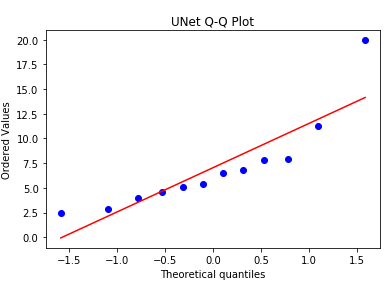}}
\qquad
\subfloat[]{\includegraphics[width = 1.9in]{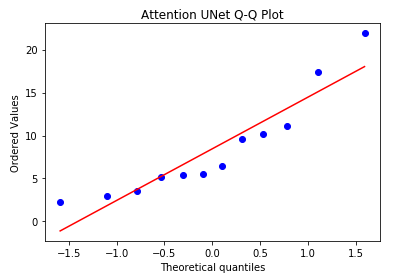}}
\qquad
\subfloat[]{\includegraphics[width = 1.9in]{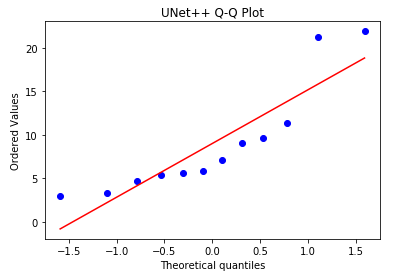}}
\caption{UNet, Attention UNet, and UNet++ Q-Q plots. The plots illustrate the deviations of the errors from a normal distribution. Figure (a) is the UNet Q-Q plot. Figure (b) is the Attention UNet Q-Q plot. Figure (c) is the UNet++ Q-Q plot.}
\label{fig:qqplots}
\end{figure}

It can be visualized from Fig. \ref{fig:qqplots} that all three models show some deviation from normality, especially at the start and end of the data. This means that our initial assumptions were correct, thus the application of a non-parametric paired statistical test is the best approach. The resulting p-values of the Wilcoxon signed rank test can be summarized in Table \ref{tab:pvalues}.

\begin{table}[!htbp]
\caption{Resulting p-value from the Wilcoxon signed rank tests between models.}
\label{tab:pvalues}
\centering
\begin{tabular}{ll}
\hline
\textbf{Comparison}       & \textbf{p-value} \\ \hline
UNet and Attention UNet   & 0.8753           \\ \hline
UNet and UNet++           & 0.2094           \\ \hline
Attention UNet and UNet++ & \textbf{0.0844}  \\ \hline
\end{tabular}
\end{table}

The highest p-value is obtained when comparing the MAPE of UNet and Attention UNet, which means that there is not enough data to reject the null hypothesis; and therefore, show a statistically significant difference to suggest one model over the other. This result matches the observed behavior where UNet would surpass Attention UNet by a high margin only on specific cases, mostly while extracting the flame height and flame area from experiments with a pipe outlet diameter of 0.015 m, during low and high mass flow rates.

The second highest p-value was obtained when comparing the MAPE of UNet and UNet++. Even if the p-value is notably lower, it remains over the established significance level, so there is not enough data to show a statistically significant difference between the models. This result goes along with the perceived errors of the specific case where UNet++ would only surpass UNet while extracting the lift-off distance from experiments with a pipe outlet diameter of 0.01275 m.

Finally, the lowest p-value is observed in the comparison between the MAPE of Attention UNet and UNet++, which is also lower than the established significance level of 0.1. This means that there is a statistically significant difference between these two models. In this case, it is recommended Attention UNet over UNet++, since the only instances where UNet++ surpassed Attention UNet happened while extracting the lift-off distance from experiments with a pipe outlet diameter of 0.01275 and 0.015 m. However, even in those cases, the difference between the errors is minimal. In the experiments with a pipe outlet diameter of 0.01275 m, the difference between errors is only 0.6 for both MAPE and RMSPE. In the experiments with a pipe outlet diameter of 0.015 m, the difference between MAPE is also 0.6, but the difference between RMSPE is only 0.1.

\section{Conclusions}
\label{sec:conclusions}

This research work explores the application of deep learning models in an alternative approach that uses the semantic segmentation of jet fires flames to extract the flame’s main geometrical attributes, relevant for fire risk analyses. To the best of our knowledge, this is the first work to present a comparative study for this approach, between traditional computer vision algorithms and certain deep learning algorithms from the state-of-the-art. The major ﬁndings of this study can be summarized as follows:

\begin{enumerate}
\item Between the analyzed traditional image processing methods and the state-of-the-art deep learning models, the best results for the segmentation of jet fires were given by the deep learning architecture UNet, along with two recent improvements on it, namely Attention UNet and UNet++. UNet obtained the best Hausdorff Distance, Attention UNet achieved the best Jaccard Index, and UNet++ took the least time to segment all 201 images from the dataset.
 
\item Compared to other traditional algorithms, Deep Learning models automatically detect the most descriptive and salient features for the problem. In contrast to the other Deep Learning architectures explored, UNet offers precise localization of the regions and a good definition of their borders. This is possible because of the skip connections that combine spatial information between its contracting and expanding path. This architecture, however, takes a significant amount of time to train and can leave high GPU memory footprints when dealing with larger images.

\item Both Attention UNet and UNet++ improve on the UNet architecture. Attention UNet reduces the number of redundant feature maps brought across by skip connections, and UNet++ bridges the semantic gap between feature maps before their concatenation. These improvements, however, increase the number of parameters, which also significantly increment the training time of the models.

\item The height of the flames is approximated from the segmentation performed by the UNet, Attention UNet, and UNet++ models. For the experiments with a pipe outlet diameter of 0.01275 m, 0.02 m, and 0.3 m, Attention UNet obtained the best results, with a maximum MAPE of 5.5\%, and a minimum of 2.3\%. For the experiment with a pipe outlet diameter of 0.015 m, UNet obtained the best result, with a MAPE of 4.5\%.

\item The area of the flames is approximated from the segmentation performed by the UNet, Attention UNet, and UNet++ models. For the experiments with a pipe outlet diameter of 0.01275 m, 0.02 m, and 0.3 m, Attention UNet obtained the best results, with a maximum MAPE of 17.3\%, and a minimum of 3.6\%. For the experiment with a pipe outlet diameter of 0.015 m, UNet obtained the best result, with a MAPE of 7.8\%.
 
\item The lift-off distances of the flames are approximated from the segmentation performed by the UNet, Attention UNet, and UNet++ models. For the experiments with a pipe outlet diameter of 0.015 m, 0.02 m, and 0.3 m, UNet obtained the best results, with a maximum MAPE of 6.5\%, and a minimum of 5.0\%. For the experiment with a pipe outlet diameter of 0.01275 m, UNet++ obtained the best result, with a MAPE of 9.6\%. This is the only instance where UNet++ outperforms the other models.

\item Among the three different geometric characteristics of jet fires analyzed in this paper, the area resulted in the highest errors from the validation experiments. This, however, can be explained by how the experimental area is usually approximated from a minimum bounding box, which usually includes areas without flame that are excluded from the segmentation’s prediction. Even so, the highest overall MAPE of 21.9\% still allows for a decent representation of the experimental area.

\item Based on the results obtained from the Wilcoxon signed rank test to compare the models’ MAPE, it can be concluded that the Attention UNet model has a better performance than the UNet++ model; since there is a statistically significant difference between them. However, the same cannot be said for the UNet model, which does not show a statistically significant difference between Attention UNet and UNet++. Further experiments and data would be necessary to show this.

\item Bearing all these results in mind, the usage of Attention UNet for general characterization could be recommended; however, further testing with UNet is also recommended for the case of lift-off distances.
 
\end{enumerate}

It is important to mention that the Deep Learning models were trained with a data set of 201 images, and using a larger data set could significantly increase the accuracy and statistically significant difference of the UNet, Attention UNet, and UNet++ architectures. Furthermore, it is important to highlight that the segmentation obtained from the current models can also be used to classify different radiation zones within the flame with their respective temperature and geometry. The information from this characterization could be used along with other models, such as a Weighted Multi Point Source model or a Computational Fluid Dynamics model. This would allow for a more thorough risk analysis of jet fires that evaluates the likelihood of direct flame impingement and determines the distribution and intensity of radiant heat that is emitted from the flame to the surrounding equipment.

As future work, other improved versions of the Deep Learning architectures could be tested and different real-time algorithms could be explored to further improve the efficiency and accuracy of the segmentation.  Additionally, the application of edge computing, smart cameras, or 8-bit quantization, could be introduced to use fewer bits for calculation and storage, which would accelerate the processing time of these algorithms. The experiment could also be redesigned for a more controlled benchmark to compare to other algorithms in the literature and future studies on the problem of fire segmentation. Taking all this into consideration, the research presented is a promising first step toward a fast and reliable risk management system for general fire incidents.

\section{Acknowledgments}
This work was supported through a scholarship for Carmina Perez-Guerrero by the Mexican National Council of Science and Technology (CONACYT). This research is part of the project 7817-2019 funded by the Jalisco State Council of Science and Technology (COECYTJAL).

\bibliography{mybibfile}

\end{document}